\documentclass{article}

\usepackage[final]{neurips_data_2021}
\usepackage[utf8]{inputenc} % allow utf-8 input
\usepackage[T1]{fontenc}    % use 8-bit T1 fonts
\usepackage{hyperref}       % hyperlinks
\usepackage{url}            % simple URL typesetting
\usepackage{booktabs}       % professional-quality tables
\usepackage{amsfonts}       % blackboard math symbols
\usepackage{nicefrac}       % compact symbols for 1/2, etc.
\usepackage{microtype}      % microtypography
\usepackage{xcolor}         % colors
\usepackage{tabulary}  
\usepackage{colortbl}
\usepackage{multirow,tabularx}
\usepackage{caption}
\usepackage{subcaption}
\usepackage{multirow}
\usepackage{comment}
\usepackage{todonotes}
\usepackage{amsmath}

\usepackage{tikz}

\newcommand*\circledd[1]{\protect\tikz[baseline=(char.base)]{
		\node[shape=circle,draw,inner sep=0pt] (char) {#1};}}

\newenvironment{titlemize}[1]{%
	\paragraph{#1}
	\begin{itemize}}
	{\end{itemize}}

\title{FEVEROUS: Fact Extraction and VERification Over Unstructured and Structured information}

\author{%
	Rami Aly, Zhijiang Guo, Michael Schlichtkrull, James Thorne, Andreas Vlachos \\
	Department of Computer Science and Technology\\
	University of Cambridge\\
	\texttt{\{rmya2,zg283,mss84,jt719,av308\}@cam.ac.uk} \\
	\And
	Christos Christodoulopoulos \\
	Amazon Alexa\\
	\texttt{chrchrs@amazon.co.uk} \\
	\AND
	Oana Cocarascu \\
	Department of Informatics \\
	King's College London \\
	\texttt{oana.cocarascu@kcl.ac.uk} \\
	\And
	Arpit Mittal \\
	Facebook\thanks{The author started working on this project whilst at Amazon.}\\
	\texttt{arpitmittal@fb.com} \\
}

\begin{document}
	
	\maketitle
	
	\begin{abstract}
		Fact verification has attracted a lot of attention in the machine learning and natural 
		language processing communities, as it is one of the key methods for detecting 
		misinformation. Existing large-scale benchmarks for this task have focused mostly on 
		textual sources, i.e.\ unstructured information, and thus ignored the wealth of 
		information available in structured formats, such as tables.
		In this paper we introduce a novel dataset and benchmark, Fact Extraction and 
		VERification Over Unstructured and Structured information (FEVEROUS), which consists of 
		87,026 verified claims. Each claim is annotated with evidence in the form of sentences 
		and/or cells from tables %and infoboxes
		in Wikipedia, as well as a label indicating whether this evidence supports, refutes, or 
		does not provide enough information to reach a verdict.
		%During the dataset construction process, 
		Furthermore, we detail our efforts to track and minimize the biases present in the 
		dataset and could be exploited by models, e.g.\ being able to predict the label without 
		using evidence.
		Finally, we develop a baseline for verifying claims against text and tables %in 
		%isolation
		which predicts both the correct evidence and verdict for 18\% of the claims.
		%achieving $0.15$ accuracy
		%when evaluated by requiring a correct prediction to have both the correct label and 
		%the correct evidence.
		%The baseline proposed returns the correct answer in terms of, both, evidence and claim 
		%verdict label for X\% of the claims. %retrieves fully supported evidence for $X\%$ and 
		%%the FEVEROUS score, which takes into account both retrieved evidence and verdict 
		%%classification, resulted in XXX.
		
	\end{abstract}
	
	\section{Introduction}

	%In the past few years there has been a lot of 
	Interest in automating fact verification has been growing as the volume of potentially 
	misleading and false claims rises \citep{2018graves}, resulting in  the development of
	%A large body of works has been developed 
	both fully automated methods (see \citet{Thorne2018AutomatedFC,Zubiaga2018DetectionAR, 
	Hardalov2021ASO} for recent surveys) as well as technologies that can assist human 
	journalists~\citep{Nakov2021AutomatedFF}.
	%There has been tremendous progress in 
	This has been enabled by the creation of datasets of appropriate scale, quality, and 
	complexity in order to develop and evaluate models for fact extraction and verification, % 
	%(see Sec.~\ref{sec:lit} for a short overview).
	e.g.\ \citet{thorneFEVERLargescaleDataset2018,augensteinMultiFCRealWorldMultiDomain2019}. 
	Most large-scale datasets focus exclusively on verification against textual evidence rather 
	than tables. Furthermore, table-based datasets, e.g.\ 
	\citet{chenTabFactLargescaleDataset2020}, assume an unrealistic setting where an evidence 
	table is provided, requiring extensions to evaluate retrieval  
	\citep{schlichtkrull2020joint}. %, and must be extended to evaluate the real-world task. %, 
	%%as they assume the table needed as evidence is provided with the claim, which is 
	%%unrealistic.
	%This is unrealistic in real-world settings.
	%Furthermore, apart from the small dataset of \citep{wang2021semeval}, they do not specify 
	%the location of the evidence in the table.
	%No previous work has investigated claims that require both structured \emph{and} 
	%unstructured evidence.
	
	\begin{figure}
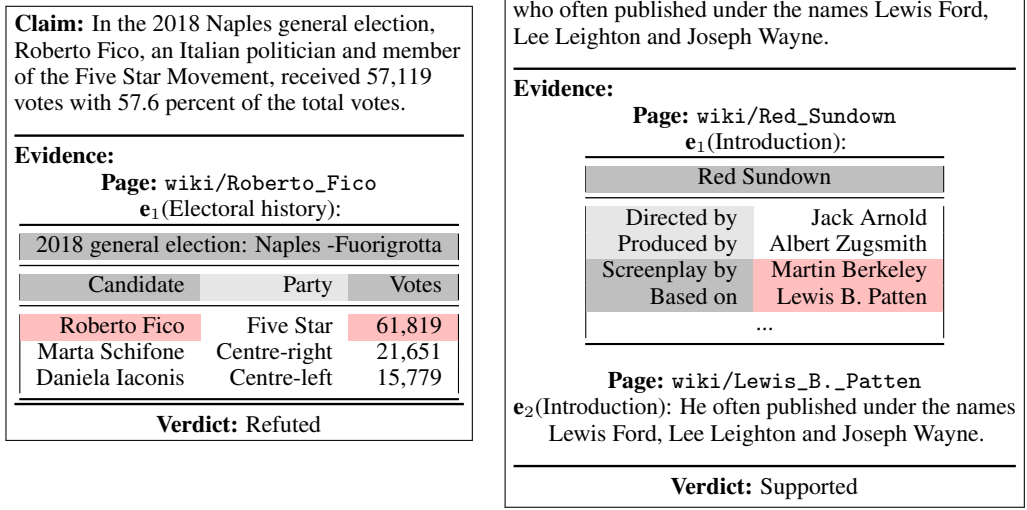

		\begin{tabular}{ c c }
			\fbox{\begin{minipage}{17em}
					\small    
					\textbf{Claim:} In the 2018 Naples general election, Roberto Fico, an 
					Italian politician and member of the Five Star Movement, received 57,119 
					votes with 57.6 percent of the total votes.\\
					\rule{\linewidth}{0.1em}
					\raggedright{\textbf{Evidence:}} \\
					\centering{\textbf{Page:} \texttt{wiki/Roberto\_Fico}}\\
					\textbf{e$_1$}(Electoral history): \centering{
						\begin{tabular}{|r r r|}
							\toprule
							\multicolumn{3}{|c|}{\cellcolor{gray!50}{2018 general election: 
							Naples -Fuorigrotta}} \\
							\midrule
							\cellcolor{gray!50}{Candidate} & \cellcolor{gray!20}{Party} & 
							\cellcolor{gray!50}{Votes} \\
							\midrule
							\cellcolor{red!25}{Roberto Fico} &	Five Star &	
							\cellcolor{red!25}{61,819} \\
							Marta Schifone &	Centre-right  &	21,651\\
							Daniela Iaconis	& Centre-left  &	15,779\\
							\bottomrule
						\end{tabular}
					}
					\rule{\linewidth}{0.1em}
					\textbf{Verdict:} Refuted\\
					%\textbf{Verification Challenge:} Combining Tables and Text\\
			\end{minipage}} & 
			
			\fbox{\begin{minipage}{19em}
					\small    
					\textbf{Claim:} Red Sundown screenplay was written by Martin Berkeley; 
					based on a story by Lewis B. Patten, who often published under the names 
					Lewis Ford, Lee Leighton and Joseph Wayne.\\
					\rule{\linewidth}{0.1em}
					\raggedright{\textbf{Evidence:}} \\
					\centering{\textbf{Page:} \texttt{wiki/Red\_Sundown}}\\
					%\textbf{S0:} \texttt{Introduction}\\
					\textbf{e$_1$}(Introduction): \centering{
						\begin{tabular}{|r r|}
							\toprule
							\multicolumn{2}{|c|}{\cellcolor{gray!50}{Red Sundown}} \\
							\midrule
							\cellcolor{gray!20}{Directed by} & Jack Arnold \\
							\cellcolor{gray!20}{Produced by} & Albert Zugsmith\\
							\cellcolor{gray!50}{Screenplay by} & \cellcolor{red!25}{Martin 
							Berkeley}\\
							\cellcolor{gray!50}{Based on} & \cellcolor{red!25}{Lewis B. 
							Patten}\\
							\multicolumn{2}{|c|}{...} \\
							\bottomrule
						\end{tabular}
					}\\
					\vspace{1em}
					\textbf{Page:} \texttt{wiki/Lewis\_B.\_Patten}\\
					%\textbf{S0:} \texttt{Introduction}\\
					\textbf{e$_2$}(Introduction): He often published under the names Lewis 
					Ford, Lee Leighton and Joseph Wayne.
					\rule{\linewidth}{0.1em}
					\textbf{Verdict:} Supported\\
					%\textbf{Verification Challenge:} Multi-hop Reasoning\\
			\end{minipage}}
		\end{tabular}
		
		\caption{
			%Example annotations (claim, evidence, and verdict) from FEVEROUS. 
			FEVEROUS sample instances. Evidence in tables is highlighted in red. Each piece of 
			evidence $e_i$ has associated context, i.e.\ page, section title(s) and the closest 
			row/column headers (highlighted in dark gray). % (i.e. for \emph{Martin Berkeley} 
			%it is \emph{Screenplay by} and \emph{Red Sundown}.)
			Left: evidence consists of two table cells refuting the claim. Right: Evidence 
			consists of two table cells and one sentence from two different pages, supporting 
			the claim.}
		%that require both unstructured and structured information to be verified %The dataset 
		%%contains shorter more simple (left) and complex claims (right).}
		\label{fig:feverous_example}
	\end{figure}
	
	In this paper, we introduce a novel dataset and benchmark, FEVEROUS: Fact Extraction and 
	VERification Over Unstructured and Structured information, consisting of claims verified 
	against Wikipedia pages and labeled as supported, refuted, or not enough information. Each 
	claim has evidence in the form of sentences and/or cells from tables in Wikipedia.
	Figure \ref{fig:feverous_example} shows two examples that illustrate the level of 
	complexity of the dataset. A claim may require a single table cell, a single sentence, or a 
	combination of multiple sentences and cells from different articles as evidence for 
	verification. %, making FEVEROUS the first to combine these two types of evidence.
	FEVEROUS contains 87,026 claims, manually constructed and verified by trained annotators. 
	Throughout the annotation process, we kept track of the two- and three-way inter-annotator 
	agreement (IAA) on random samples %using 6\% and 3\% of the dataset respectively,
	with the IAA kappa $\kappa$ being $0.65$ for both. %and the three-way IAA was $0.65$.
	Furthermore, we checked against dataset annotation biases, such as words present in the 
	claims that indicate the label irrespective of evidence~\citep{Schuster2019TowardsDF}, and 
	ensured these are minimised.

\begin{figure}[h]
		\centering
		\includegraphics[scale=0.59]{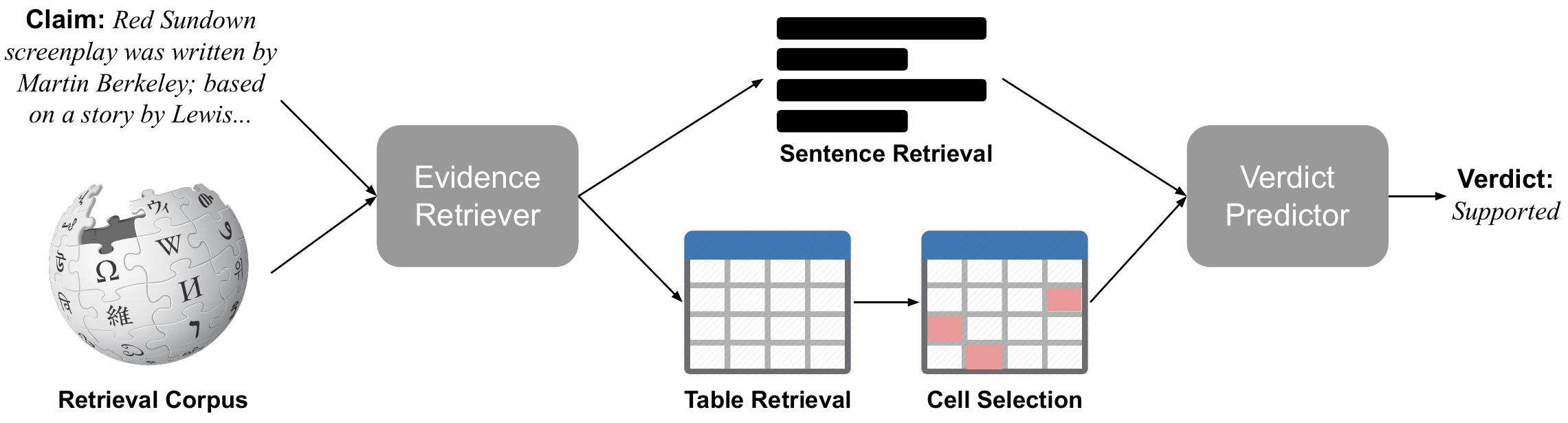}
		\caption{The pipeline of our FEVEROUS baseline.}
		\label{fig:baseline}
	\end{figure}

	We also develop a baseline approach to assess the feasibility of the task defined by 
	FEVEROUS, 
	%. The pipeline of our model is 
	shown in Figure \ref{fig:baseline}.
	We employ a combination of entity matching and TF-IDF %the Dense Passage 
	%Retriever~\citep{karpukhin-etal-2020-dense} 
	to extract the most relevant sentences and tables to retrieve %that may form
	the evidence, followed by a cell extraction model that returns relevant cells from tables 
	by linearizing them and treating the extraction as a sequence labelling task. A RoBERTa 
	classifier pre-trained on multiple NLI datasets %including 
	%FEVER~\citep{nie-etal-2020-adversarial}, 
	predicts the veracity of the claim using the retrieved evidence and its context. This 
	baseline substantially outperforms the sentence-only and table-only baselines.
	The proposed baseline predicts correctly both the evidence and the verdict label for 18\% 
	of the claims. The retrieval module itself fully covers 28\%  of a claims evidence. %Our 
	%error analysis shows that our method ... 
	FEVEROUS is the first large-scale verification dataset that focuses on sentences, tables, 
	and the combination of the two, and we hope it will stimulate further progress in fact 
	extraction and verification  % as most previous large-scale datasets and shared tasks 
	%relying on them were focused exclusively on textual sources. 
	%To this end, we are currently organising a shared task, 
	and is publicly available online: \url{https://fever.ai/dataset/feverous.html}. %), along 
	%with links to the dataset and the code of this paper.

	\section{Literature Review}
	\label{sec:lit}
	Datasets for fact verification often rely on real-world claims from fact-checking websites 
	such as  PolitiFact. % (\url{https://www.politifact.com/}). 
	%%%citet{vlachosFactCheckingTask2014} introduced a small dataset of 106 claims with curated 
	%%annotations justifying decisions and providing sources. 
	%%
	For such claims, the cost of constructing fine-grained evidence sets can be prohibitive. 
	Datasets therefore either leave out evidence and justifications 
	entirely~\citep{wangLiarLiarPants2017a}, rely on search engines which risk including 
	irrelevant or misleading 
	evidence~\citep{Popat2016CredibilityAO,Baly2018IntegratingSD,augensteinMultiFCRealWorldMultiDomain2019},
	 or bypass the retrieval challenge entirely by extracting evidence directly from the fact 
	checking 
	articles~\citep{Alhindi2018WhereIY,hanselowskiRichlyAnnotatedCorpus2019,kotonya2020explainable}
	 or scientific literature~\citep{Wadden2020FactOF}.
	%\citet{wangLiarLiarPants2017a} proposed a dataset with 12.8k claims, but, whilst large 
	%enough to train machine learning models, the justifications and evidence were left out as 
	%the construction of appropriate evidence sets for such real-world datasets is expensive.
	%%
	%Other efforts have used results from search engines instead of curated 
	%sources~\citep{Popat2016CredibilityAO,Baly2018IntegratingSD,augensteinMultiFCRealWorldMultiDomain2019},
	% which addresses the scaling problem but risks including irrelevant or misleading 
	%evidence. 
	%Several recent papers have taken a different approach, extracting evidence directly from 
	%fact checking articles~\citep{Alhindi2018WhereIY,hanselowskiRichlyAnnotatedCorpus2019}. 
	%This allows the construction of large-scale datasets with appropriate evidence, but 
	%bypasses the retrieval challenge and assumes that appropriate evidence has already been 
	%collected.
	
	The cost of curating evidence sets for real-world claims can be circumvented by creating 
	artificial claims. \citet{thorneFEVERLargescaleDataset2018} introduced FEVER, a large-scale 
	dataset of 185,445 claims constructed by annotators based on Wikipedia articles. 
	% This dataset was later extended with additional adversarial 
	%examples~\citep{Thorne2019TheFS}. 
	This annotation strategy %behind FEVER
	was adopted  %subsequently used
	to construct a similar dataset for Danish~\citep{danfever2021}, and adapted for real-world 
	climate change-related claims~\citep{diggelmannCLIMATEFEVERDatasetVerification2021}. 
	\citet{jiangHoVerDatasetManyHop2020} extended this methodology to create a dataset of 26k 
	claims requiring multi-hop reasoning. 
	Other annotation strategies include % efforts
	%have sought to expand fact verification to new languages and domains. 
	\citet{Khouja2020StancePA} who introduced a dataset of Arabic claims generating supported 
	and unsupported claims based on news articles. 
	
	% and \citet{Wadden2020FactOF} who proposed a dataset for fact verification of scientific 
	%statements using article abstracts as evidence, while \citet{kotonya2020explainable} 
	%introduced a similar dataset for the medical domain.
	%%
	% Such synthetic datasets are often constructed using crowd-sourcing mechanisms. Prior work 
	%showed that artificial claims contained idiosyncrasies that can be 
	%exploited~\citep{Schuster2019TowardsDF}. Adversarial evaluation has been used to explore 
	%the biases behind FEVER and additional adversarial examples have been 
	%incorporated~\citep{Thorne2019TheFS}. To alleviate such biases, recent synthetic datasets 
	%either introduced a multi-player game to collect adversarial data~\citep{foolme2021}, or 
	%leveraged context-sensitive inference to improve the robustness to adversarial 
	%inputs~\citep{vitaminc2021}. 
	
	The datasets discussed so far have primarily focused on unstructured text as 
	%, excluding everything except textual 
	evidence during the annotation process. %~\citep{thorneFEVERLargescaleDataset2018}. 
	There is currently a small number of datasets %Recent efforts %have been taken to 
	%have developed datasets 
	that rely on structured information, primarily tables. 
	TabFact~\citep{chenTabFactLargescaleDataset2020} and 
	InfoTABS~\citep{guptaINFOTABSInferenceTables2020} contain artificial claims to be verified 
	on the basis of Wikipedia tables and infoboxes respectively, while 
	SEM-TAB-FACTS~\citep{wang2021semeval} requires verification on the basis of tables from 
	scientific articles. The latter is the only to also specify the location of the evidence in 
	a table. %Furthermore, apart from the small dataset of \citep{wang2021semeval}, they do not 
	%specify the location of the evidence in the table.
	%\citet{chenTabFactLargescaleDataset2020} introduced TabFact, a dataset of 118k 
	%human-produced claims over tables collected from Wikipedia. Similarly, 
	%\citet{guptaINFOTABSInferenceTables2020} proposed InfoTABS, a dataset of 23,738 claims 
	%constructed based on Wikipedia infoboxes. More recently, \citet{wang2021semeval} 
	%constructed SEM-TAB-FACTS, consisting of 3,764 claims and 957 tables from scientific 
	%articles.  %In addition to verifying the claim, each cell in the table needs to be 
	%%selected 
	%%as evidence.
	%Contrary to our proposed dataset, these corpora either do not require models to perform 
	%evidence retrieval, and/or include \textit{only} structured evidence.
	Our proposed dataset is the first which considers both structured \emph{and} unstructured 
	evidence for verification, while 
	%and which contains claims that require models to include both for verification.
	%Furthermore, our dataset is the first to 
	explicitly requiring the retrieval of evidence. % as well as claim verification for 
	%structured data.
	
	In the related field of question answering~\citep{BOUZIANE2015366}, recent work also 
	considered finding answers over both tables and text. \citet{chen-etal-2020-hybridqa} 
	proposed HybridQA, a dataset consisting of multi-hop questions constructed by using 
	Wikipedia tables and the introductory section of linked entities in the table, however, 
	their dataset assumes the table as part of the input. Based on HybridQA, 
	\citet{chen2021open} further required systems to retrieve relevant tables and texts by 
	decontextualizing questions of HybridQA and adding additional questions to remove potential 
	biases, resulting in a total of 45K question-answer pairs. The NaturalQA 
	dataset~\citep{47761} is substantially larger (about 300K) with some questions requiring to 
	retrieve answers from tables (17\%). However, these tables are predominantly infoboxes and 
	rarely require systems to combine information from both text and tables. 
	
	%While the NaturalQA dataset is substantially larger and a \citep{47761} proposed a 
	%substentially larger datase  dataset of  decontextualize HybridQA considers the retrieval
	%multiple free-form corpora linked with the entities in the table
	
	%\citet{vitaminc2021}
	
	%Importance of evidence + justification?
	
	\section{FEVEROUS Dataset and Benchmark}
	
	%The aim of the FEVEROUS task is to assess
	In FEVEROUS the goal is to determine the veracity of a claim $c$ by: \textit{i)} retrieving 
	a set of evidence pieces $E$ which can be either a sentence or a table cell, and 
	\textit{ii)} assigning a label $y\in\{\text{Supports}, \text{Refutes}, \text{Not Enough 
	Info}\}$. %$E$ can be any combination of different types of evidence. 
	The source of evidence is derived from the English Wikipedia (excluding pages and sections 
	flagged to require addition references or citations), and consists of sentences and tables 
	obtained as follows: 
	%The evidence to verify the claim has to be extracted from a retrieval corpus,  for 
	%evidence retrieval 
	%A table $t\in \mathbb{B}_T$ consists of cells $c_{i,j}$ with $i$ and $j$ specifying the 
	%row and column, respectively, as well as in some cases a caption $q$.  A list $l\in 
	%\mathbb{B}_L$ is a collection of list items $i_{i,j}$, with $i$, $j$ indicating the list 
	%depth and position index, respectively. A piece of evidence $e \in E$ is thus be defined 
	%to 
	%be $e \in \mathbb{B}_S \cup \{c_{i,j} | c_{i,j} \in t \land t \in \mathbb{B}_T \} \cup \{q 
	%| q \in t \land t \in \mathbb{B}_T \} \cup \{l_{i,j} | l_{i,j} \in l \land l \in 
	%\mathbb{B}_L \}$. An annotation can contain up to three evidence sets, in case different 
	%pieces of evidence lead to the same verdict independently. All elements of the retrieval 
	%corprus maintain hyperlinks to other Wikipedia articles. 
	
	\textbf{Sentence.} Any sentence from a Wikipedia article's text as well as special 
	Wikipedia phrases referring to other articles (e.g. \emph{See also: ...}, \emph{X redirects 
	here. For other uses, see ...}). % indicate a sentence.
	
	\textbf{Table.} A table consists of cells $c_{i,j}$, where $i$ and $j$ specify the row and 
	column, respectively, and a caption $q$. Both cells and captions can take various formats 
	like a single word, number or symbol, phrases, and entire sentences. %as well  into tables, 
	%which are  categorize Tables into \emph{Infobox} and \emph{Wikitable}, as the former have 
	%distinct properties from other tables on Wikipedia. Wikipedia tables are diverse, with a 
	%varying degree of complexity in their structure. 
	% 
	%The contents of a cell $c_{i,j}$ may include numbers, words, symbols, phrases, entire 
	%sentences. 
	%%
	In most datasets (e.g.~\citet{chenTabFactLargescaleDataset2020}), headers are restricted to 
	the first row of a table. However, tables in Wikipedia can have a more complex structure, 
	including multi-level headers (see Figure \ref{fig:feverous_example} for an example). 
	FEVEROUS maintains the diversity of Wikipedia tables, only filtering out those with 
	formatting errors. %Cell headers can be located anywhere in the tables as well, and not 
	%only in the first row/column of a table, such as in 
	%\citep{chenTabFactLargescaleDataset2020}. 
	%%
	For the purposes of annotation, a table caption $q$ is considered to be a table cell and 
	needs to be selected explicitly if it contains information relevant to the claim. %can take 
	%various formats, e.g., a single word, phrase, entire sentence. % A table caption $q$ is 
	%%considered to be a table cell as well and can thus be of any format as well, ranging from 
	%%a 
	%%single word, a phrase, to an entire sentence(s). 
	%%
	We also include Wikipedia infoboxes as tables, as well as lists.
	%In addition to Wikitables, we include infoboxes as tables. % We considered to be a cell in 
	%%the context of the paper.
	%%
	We consider the latter 
	%We also include Wikipedia Lists in the retrieval corpus and consider them 
	to be special tables where the number of items in the list yields the number of columns and 
	the number of nested lists yields the number of rows. For example, $c_{1,5}$ represents the 
	item of a nested list at depth $1$ found at the fifth position of the main list. %List 
	%tables are further categorized as being \emph{itemized} or \emph{enumerated}. %We also 
	
	The evidence retrieval in FEVEROUS considers the entirety of a Wikipedia article and thus 
	the evidence can be located in any section of the article except the reference sections. 
	The order between all elements in an article is maintained. %and related work that only 
	%considers the introductory section of an article. 
	%To attribute for the position of elements in an article, each element (and, thus, piece of 
	%evidence) has associated \emph{context}. 
	We associate each candidate piece of evidence with its \emph{context}, 
	%The context for an element is the 
	which consists of the article's title and section titles, including the sub-sections the 
	element is located in. For table cells, we also include the nearest row and column headers; 
	if the element just before the nearest row/column is also a header, then it will be 
	included in the context.
	%are additionally considered as the context of the cell. 
	Context adds relevant information to understand a piece of evidence, but it is not 
	considered a piece of evidence by itself. 
	%The retrieval corpus consists of 5.4M articles with 95.6M sentences and 11.8M tables (5.8M 
	%tables and 6M lists). %5.8 million tables, and 6 million lists. 
	Sentences and cells maintain their hyperlinks to other Wikipedia articles, if present. 
	%Articles and sections flagged by Wikipedia to require additional references or citations 
	%were excluded. % from the retrieval corpus.  %For a given piece of evidence, there is 
	%%%associated context that can be used.with an ordering  with $e\in\mathbb{B}_S \cup 
	%%%\mathbb{B}_T \cup \mathbb{B}_L$.
	
	Quantitative characteristics of FEVEROUS and most related fact-checking datasets (i.e.\ 
	FEVER, TabFact, and 
	Sem-Tab-Facts) are shown in Table \ref{tab:dataset-statistics}.
	As seen, the average claim of FEVEROUS is more than twice as long as the other datasets.
	%The claim length correlates with the number of required evidence pieces, requiring
	On average $1.4$ sentences and $3.3$ cells (or $0.8$ Tables) are required as evidence per 
	sample, higher than both FEVER and Sem-Tab-Facts \emph{combined}. Looking into the evidence 
	sets by type, we note that FEVEROUS is balanced, having almost an equal amount of instances 
	containing, either exclusively text, tables, or both as evidence. Regarding the veracity 
	labels, FEVEROUS is roughly balanced in terms of supported (56\%) and refuted claims 
	(39\%), with only about 5\% of claims being NotEnoughInfo.
	
	\begin{table}[ht!]
		\centering
		\caption{Quantitative characteristics of FEVEROUS compared to related datasets. Claim 
		length is reported in tokens. \emph{Avg. Evidence} is the average number of evidence 
		pieces per claim in a dataset, while \emph{Evidence Sets by type} reports the number of 
		unique evidence sets by type. For FEVEROUS, \emph{combined} measures the number of 
		annotations that require evidence from both tables and sentences. The \emph{Evidence 
		Sets} can be used as \emph{Evidence Source} for SEM-TAB-FACTS and TabFact, as explored 
		by \citet{schlichtkrull2020joint} for the latter.
		}\label{tab:dataset-statistics}
		\resizebox{\columnwidth}{!}{%
			\begin{tabular}{c|c c c c}
				\toprule
				\textbf{Statistic} &  \textbf{FEVEROUS} & \textbf{FEVER} & \textbf{TabFact} & 
				\textbf{SEM-TAB-FACTS} \\
				\midrule
				\midrule
				Total Claims & 87,026  & 185,445 & 117,854  & 5,715 \\
				\midrule
				Avg. Claim Length & 25.3 &  9.4 & 13.8 & 11.4 \\
				\midrule
				\multirow{2}{*}{Avg. Evidence} & 1.4 sentences, 3.3 cells & 1.2 %268,396 
				%1.45
				sentences &  1 table  & 1.1 cells    \\ %6505 cells in sem-tab-facts, 6.6 cells 
				%over only dev-test
				& (0.8 tables) & & & (1 table)\\
				\midrule
				\multirow{2}{*}{Evidence Sets by Type} & 34,963 sentences, 28,760 tables, &  
				296,712 sets  & 16,573 tables & 1,085 tables\\ 
				&  24,667 combined &  &  &   \\
				\midrule
				%Number Evidence & $4.55$ & XX & 1 Table & 1 Table,  \\
				\multirow{2}{*}{Size of Evidence Source} & 95.6M sentences, & 25.1M sentences, 
				& 16,573 tables & 1,085 tables \\ 
				& 11.8M tables &  & & \\ % Fever: 5.4M documents
				\midrule
				Veracity Labels & 49,115 Supported, & 93,367 Supported, & 63,723 Supported, & 
				3,342 Supported, \\
				&  33,669 Refuted, & 43,107 Refuted, & 54,131 Refuted & 2,149 Refuted, \\
				& 4,242 NEI &  48,973 NEI &  & 224 Unknown\\
				\bottomrule
		\end{tabular}}
		
	\end{table}

	\subsection{Dataset Annotation}
	%Similar to FEVER, 
	Each claim in the FEVEROUS dataset was constructed in two stages: \textit{1)} %extraction 
	%and
	claim generation %of claims
	based on a Wikipedia article, \textit{2)} retrieval of evidence from Wikipedia and 
	selection of the appropriate verdict label, i.e.\ claim verification. Each claim is 
	verified by a different annotator than the one who generated it to ensure the verification 
	is done without knowledge of the label or the evidence.
	%ensuring that there is no overlap between annotators of claim generation and those of 
	%verification.
	%%
	A dedicated interface built on top of Wikipedia's underlying software, Mediawiki 
	(\url{https://www.mediawiki.org/wiki/MediaWiki}) to give annotators a natural and intuitive 
	environment for searching and retrieving relevant information. The ElasticSearch engine, in 
	particular the CirrusSearch Extension, allowed for more advanced search expressions with 
	well-defined operators and hyperlink navigation, as well as a custom built page search 
	functionality, enabling annotators to search for specific phrases in an article. 
	This interface allows for a %more
	diverse generation of claims, as annotators can easily combine information from multiple 
	sections or pages. See the supplementary material for screenshots of the interface and 
	examples of its use. 
	We logged all of the annotators' interactions with the platform (e.g.\ search terms 
	entered, hyperlinks clicked) and include them in the FEVEROUS dataset, as this information 
	could be used to refine retrieval models with additional information on intermediate 
	searches and search paths that led to relevant pages. 
	%The retrieval of relevant evidence is more difficult compared to annotation tasks such as 
	%classification given that information can be found in any section of an article, thus, it 
	%was necessary to build a sophisticated annotation environment. Additionally, the interface 
	%allows for a more diverse generation of claims, as annotators can easily combine 
	%information from multiple sections or pages of Wikipedia articles. %On top of the 
	%%Mediawiki 
	%%environment, we added a task-specific interface for each annotation task.
	
	\subsubsection{Claim Generation}
	\label{sec:claim-generation}
	
	To generate a claim, annotators were given a \emph{highlight} of either four consecutive 
	sentences or a table, located anywhere in a Wikipedia page; each page is used only once, 
	i.e.\ only one set of claims is generated per page, to prevent the generation of claims 
	that are too similar.
	This allowed us to control the information that annotators used and consequently the 
	distribution of topics in the claims. Sentence highlights are restricted to sentences that 
	have at least $5$ tokens, whereas table highlights must have at least $5$ rows and at most 
	$50$ rows. These bounds have been chosen based on previous work, with the rows upper bound 
	being equal to TabFact's and the lower bounds being equal to HybridQA's. While TabFact does 
	not use lower bounds, we noticed that it is infeasible to construct more complicated claims 
	from tables with fewer than $5$ rows.

	% These bounds have been selected using related datasets, specifically TabFacts for the 
	%upper bounds 
	%%
	The sentence versus table highlights ratio is 1:2. % The ratio of highlights is 1:2 
	%sentence versus table highlights. 
	Annotators had the option to skip highlights if the sentences/tables had formatting issues 
	or if the content 
	% contradicted
	enable the creation of %for did not satisfy the requirements set creating
	verifiable, unambiguous, and objective claims (see supplementary material for the full list 
	of requirements).
	For each highlight, annotators were asked to write three different claims with the 
	specifications described below, each claim being a factual and well-formed sentence. % 
	%While we enforced for the first claim an equal balance between supported and refuted 
	%claims,  specified  Annotators were instructed for the first claim whether the claim  for 
	%the first claim whether it the first clai
	%%
	% We set out a list of requirements to ensure that 
	% The claims are, amongst others, verifiable, unambiguous, and not subjective (the full 
	%list of requirements can be found in the supplementary material).
	
	\textbf{Claim using highlight only} (Type I). This type of claim must use information 
	exclusively from the highlighted table/sentences and their context (page/section titles or 
	headers). For sentence highlights we did not allow claims to be paraphrases of one of the 
	highlighted sentences, but to combine information from the four highlighted sentences 
	instead. For claims based on a table highlight, annotators were asked to combine 
	information from multiple cells if possible,
	using comparisons, filters, arithmetic and min-max operations. Only for the first claim we 
	also specified the veracity of the generated claim, enforcing an equal balance between 
	supported and refuted claims. This decision was motivated by the observation that 
	annotators have a strong tendency to write supported claims as these are more natural to 
	generate. For both Type II and III claims, annotators could freely decide to create either 
	supported, refuted, or NEI claims, as long as they adhere to the claim requirements.

	\textbf{Claim beyond the highlight} (Type II). This type of claim must be based on the 
	highlight, but must also include information beyond it. Annotators could either modify the 
	first claim they generated or create an unrelated claim that still included information 
	from the highlight. 
	Furthermore, we enforced with equal probability whether the claim had to use information 
	exclusively from the same page or from multiple pages.
	% We further added a flag whether the claim had to use information exclusively from the 
	%same page or multiple pages. 
	For the latter, annotators were allowed to navigate Wikipedia using the search engine and 
	page search tools %navigation tools as 
	previously described.
	
	\textbf{Mutated Claim} (Type III). We asked annotators to modify one of the two claims 
	previously generated using one of the following `mutations': \emph{More Specific}, 
	\emph{Generalization}, \emph{Negation}, \emph{Paraphrasing}, or \emph{Entity Substitution}, 
	with probabilities $0.15$, $0.15$, $0.3$, $0.1$, $0.3$, respectively. These mutations are 
	similar, but less restrictive than those used in FEVER (see supplementary material). 
	Annotators were allowed to navigate Wikipedia freely to extract information for generating 
	this claim. 
	
	For each generated claim, the annotators were also asked to specify the main challenge they 
	expect a fact-checker would face when verifying that claim, selecting one out of six 
	challenge categories: claims that require evidence from two or more sections or articles 
	(\textit{Multi-hop Reasoning}), combination of structured and unstructured evidence 
	(\textit{Combining Tables and Text}),  reasoning that involves numbers or arithmetic 
	operations (\textit{Numerical Reasoning}), disambiguation of entities in claims 
	(\textit{Entity Disambiguation}), requiring search terms beyond entities mentioned in claim 
	(\textit{Search terms not in claim}), and \textit{Other}.

	\subsubsection{Claim Verification}
	Given a claim from the previous annotation step, annotators were asked to retrieve evidence 
	and determine whether a claim is supported or refuted by evidence found on Wikipedia. Each 
	annotation may contain up to three possibly partially overlapping evidence sets, and each 
	set %different pieces of evidence 
	leads to the same verdict independently. %Evidence sets may consist of partially 
	%overlapping evidence. 
	For supported claims, every piece of information has to be verified by evidence, whereas 
	for refuted claims, the evidence only needs to be sufficient to refute one part of the 
	claim. %, i.e.\ annotations for refuted claims do not need to contain evidence for parts of 
	%the claim that are supported.  
	If the verification of a claim requires to include every entry in a table row/column (e.g.\ 
	claims with universal quantification such as ``highest number of gold medals out of all 
	countries''), each cell of that row/column is highlighted. In some cases, a claim can be 
	considered unverifiable (Not Enough Information; NEI) if not enough information can be 
	found on Wikipedia to arrive at one of the two other verdicts. 
	In contrast to FEVER dataset, we also require annotated evidence for NEI claims capturing
	%. In this case, the annotated evidence is 
	the most relevant information to verification of the claim, even if that was not possible. 
	This ensures that all verdict labels are equally difficult to predict correctly, as they 
	all require evidence.
	
	%The claim verification interface is shown in Figure \ref{fig:interface_overview}. 
	Starting from the Wikipedia search page, annotators were allowed to navigate freely through 
	Wikipedia to find relevant evidence. They were also shown the associated context of the 
	selected evidence in order to assess whether the evidence is sufficient on its own given 
	the context or whether additional evidence needs to be highlighted. %to provide additional 
	%information.
	Before submitting, annotators were shown a confirmation screen with the highlighted 
	evidence, the context, and the selected verdict, to ensure that all required evidence has 
	been highlighted and that they are confident in the label they have selected. 
	
	%\paragraph{Refuting claims:}
	While we require information to be explicitly mentioned in the evidence
	in order to support a claim, we noticed that requesting the same for refuting claims would 
	lead to counter-intuitive verdict labels. For example, ``Shakira is Canadian'' would be 
	labelled as NEI when we consider the evidence ``Shakira is a Colombian singer, songwriter, 
	dancer, and record producer'' \emph{and} no mention of Shakira having a second nationality 
	or any other relation to Canada. A NEI verdict in this case is rather forced and unnatural, 
	as there is no reason to believe that Shakira could be Canadian given the Wikipedia 
	article. To address these cases, we added a guideline question ``Would you consider 
	yourself misled by the claim given the evidence you found?'', so that, if answered yes (as 
	in the above example), claims are labelled as Refuted, otherwise they are labelled NEI. 
	This label rationale is different from FEVER for which explicit evidence is required to 
	refute a claim. While it could be argued that, %albeit more intuitive, 
	our approach to labelling claims leaves potentially more room for ambiguity as the decision 
	partially depends on what the annotator expects to find on a Wikipedia page and whether a 
	claim adheres to the Grice's Maxim of Quantity (being as informative as possible, giving as 
	much information as needed), our quality assessment shows that verdict agreement is very 
	high when the annotated evidence is identical  %with a kappa score of above 0.9
	(see  Section \ref{sec:quality-control}).
	
	After finishing the verification of the given claim, annotators then had to specify the 
	main challenge for verifying it, using the same six challenge categories as for the 
	challenge prediction in section~\ref{sec:claim-generation}. %selecting one out of six 
	%challenge categories: claims that require evidence from two or more sections or articles 
	%(\textit{Multi-hop Reasoning}), combination of structured and unstructured evidence 
	%(\textit{Combining Tables and Text}),  reasoning that involves numbers or arithmetic 
	%operations (\textit{Numerical Reasoning}), disambiguation of entities in claims 
	%(\textit{Entity Disambiguation}), search terms beyond entities mentioned in claim 
	%(\textit{Search terms not in claim}), and \textit{Other}.
	Examples and quantitative characteristics %and an overview on when expected and actual 
	%challenges
	on expected and actual challenges can be found in the supplementary material. %align or 
	
	\subsection{Quality Control}
	\label{sec:quality-control}
	
	\paragraph{Annotators:} Annotators were hired through an external contractor. A total of 57 
	and 54 annotators were employed for the claim generation and claim verification stages 
	respectively.%, ensuring each claim is generated and verified by different annotators. % no 
	%%overlap between the two groups.
	%We ensured that there is no overlap between annotators of claim generation and those of 
	%verification to minimise biases that could transfer from one task to the other.  
	The annotations were supervised by three project managers as well as the authors of this 
	paper. For claim generation, half of the annotators were native US-English speakers, while 
	the other half were \emph{language-aware} (an upper education degree in a language-related 
	subject). 
	English speakers from the Philippines, whereas the evidence annotators had to be 
	language-aware native US-English speakers. 
	The annotator candidates were screened internally by the external contractor to assess 
	their suitability for this task. The screening followed a two-stage process. First, the 
	candidates' English proficiency was assessed through grammatical, spelling, and fluency 
	tests. Second, the candidates were asked to give a sentence-long summary for a given 
	paragraph that they would then be asked to mutate by means of negation or entity 
	substitution, similarly to Section \ref{sec:claim-generation}. %and was tailored closer to 
	%the FEVEROUS task.
	The same screening procedure was used for both tasks, with the difference that the minimum 
	score was set higher for the claim verification part. Details on the annotator demographics 
	can be found in the supplementary material. %Moreover, for task ii) applicants mu For task 
	%i), half of the annotators were native US-English while the other half were native 
	%Philippine English speakers with a total of X annotators. For Task ii) only native 
	%US-English speakers were employed % with XXXXX.

	\paragraph{Calibration:} 
	Due to the complexity of the annotation, we used a two-phase calibration procedure for 
	training and selecting annotators. For this purpose, highlights with generated claims 
	annotated with evidence and verdicts were created by the authors to cover a variety of 
	scenarios. While the first calibration phase aimed at familiarizing the annotators with the 
	task, the second phase contained more difficult examples and special cases. Annotators had 
	to annotate a total of ten highlights/claims in each calibration phase.  Annotations for 
	claim generation were graded by the project managers in a binary fashion, i.e.\ whether a 
	claim adheres to the guideline requirements or not and whether the expected challenge is 
	appropriate. For claim verification they were graded using the gold annotations by the 
	authors using
	%the authors compiled a set of gold annotations for each claim so that annotations could be 
	%automatically %scored. The computed scores included 
	label accuracy, evidence precision/recall (see Section \ref{sec:evaluation}), the number of 
	complete evidence sets, and selected verification challenge.
	Before continuing with the second calibration phase, annotators had to review the scores 
	and feedback they received %regarding their annotations 
	in the first phase.
	Based on the scores in both phases, the project managers approved or rejected each 
	annotator, with an approval rate of $40\%$ for claim generation and $54\%$ for claim 
	verification, with a total of $141$ and $100$ claim generation and verification candidates, 
	respectively. % $100$ resulting in $84$ out of $141$ claim generation and $46$ out of $100$ 
	%claim verification candidates being rejected.
	
	\paragraph{Quality Assurance:} Generated claims were quality checked by claim verification 
	annotators who had to report those that did not adhere to the claim requirements, 
	specifying the reason(s). %by selecting one or multiple predefined reasons and/or 
	%formulating an individual reason. 
	$2534$ claims were reported with the most common reason being \emph{Ungrammatical, spelling 
	mistakes, typographical errors}, followed by \emph{Cannot be verified using any publicly 
	available information}. %Moreover, only 13\% and 7\% of not refuted claims (as refuted 
	
	Around 10\% of the claim verification annotations ($8474$ samples) were used for quality 
	assurance. %by using random sampling.
	We measured two-way IAA using 66\% of these samples, and three-way agreement with the 
	remaining 33\%. The samples were selected randomly proportionally to the number of 
	annotations by each annotator. The $\kappa$ over the verdict label was $0.65$ both for 
	two-way and three-way agreement. Duplicate annotations (and hence disagreements) due to 
	measuring IAA are not considered for the dataset itself.
	These IAA scores are slightly lower than the ones reported for FEVER dataset ($0.68$), 
	however the complexity of FEVEROUS is greater as entire Wikipedia pages with both text and 
	tables are considered as evidence instead of only sentences from the introductory sections.
	%considering the higher complexity of FEVEROUS, our numbers appear to be well in range 
	%across all samples considered for QA.
	TabFact has an agreement of $0.75$, yet in TabFact the (correct) evidence is given to the 
	annotators. If we only look at claims where annotators chose identical evidence, verdict 
	agreement in FEVEROUS is very high ($0.92$), showing that most disagreement is caused by 
	the evidence annotation. %indicating almost perfect agreement once evidence was retrieved. 
	%%% indicating  H, in TabFact considering the higher complexity of FEVEROUS, our numbers 
	%%appear to be well in range.
	Pairs of annotators annotated the same evidence for 42\% of the claims and partially 
	overlapping evidence of at least 70\% for 74\% of them. In 27\% of the claims the evidence 
	of one annotator is a proper subsets of another, indicating that in some cases evidence 
	might provide more information than required, e.g. identical information that should have 
	been assigned to two different evidence set. %in around 27\% of pairs 
	
	Further analysing the cases of disagreement, we observe that in a third of two-way IAA 
	disagreement cases one annotator selected NEI, which is disproportionately high considering 
	NEI claims make up only 5\% of the dataset, again indicating that the retrieval of evidence 
	is a crucial part of the verification task. For the other two-thirds, when annotators 
	selected opposite veracity labels we identified four sources of disagreement: (i) numerical 
	claims that require counting a large number of cells, so small counting errors lead to 
	opposing labels (ii) long claims with a refutable detail that had been overlooked and hence 
	classified erroneously (iii) not finding evidence that refutes/supports a claim due to 
	relevant pages being difficult to find (e.g.\ when the article's title does not appear in 
	the claim) (iv) accidental errors/noise, likely caused by the complexity of the task. 
	Looking into the IAA between every annotator pair shows an overall consistent annotation 
	quality with a standard deviation of $0.07$ and a total of $10$ annotators with an average 
	IAA of below $0.60$, and $8$ being higher than $0.70$. % with the minimum and maximum IAA 
	
	\paragraph{Dataset Artifacts \& Biases:}
	To counteract possible dataset artifacts, we measured the association between several 
	variables, using normalized PMI throughout the annotation process. We found that %Between 
	%most variables,  %or between words in the evidence words) 
	no strong co-occurrence was measured between the verdict and the words in the claim, 
	indicating that no claim-only bias \citep{Schuster2019TowardsDF} is present in the dataset.
	We observed the following correlations: an evidence table/sentence being the first element 
	on a page with supported verdict (nPMI=0.14) and after position $20$ with NEI verdict 
	(nPMI=0.09); words 'which/who' with Claim Type II as well as mutation type \emph{More 
	specific} and \emph{Entity Substitution} (nPMI=0.07); Claim Type II with supported verdict 
	(nPMI=0.17) and Claim Type III with refuted label (nPMI=0.23). The latter can most likely 
	be attributed to the \emph{Negation} and \emph{Entity substitution} mutations. Since we do 
	not release the claim-type correspondence, the association of words with claim types and 
	mutations is not of concern. 
	
	% To further ensure that no hidden artifacts allow reaching the right verdict without the 
	%use of the right evidence, 
	We also developed a \textbf{claim-only baseline}, which uses the claim as input and 
	predicts the verdict label. We opted to fine-tune a pre-trained BERT model 
	\citep{devlin-etal-2019-bert} with a linear layer on top and measured its accuracy using 
	5-fold cross-validation. This claim-only baseline achieves $0.58$ label accuracy, compared 
	to the majority baseline being $0.56$. Compared to FEVER where a similar claim-only 
	baseline achieves a score of about $0.62$ over a majority baseline of $0.33$ 
	\citep{Schuster2019TowardsDF}, the artefacts in FEVEROUS appear to be minimal in this 
	respect.
	Regarding the position of the evidence, we observed that cell evidence tends to be located 
	in the first half of a table. For smaller tables, evidence is more evenly distributed 
	across rows. Moreover, a substantial amount of claims require using entire columns as 
	evidence, and thus the later parts of a table as well. %, so in these cases the entire 
	
	Finally, we trained a \textbf{claim-only evidence type model} to predict whether a claim 
	requires as evidence sentences, cells, or a combination of both evidence types. The model 
	and experimental setup were identical to the one used to assess claim-only bias. The model 
	achieved $0.62$ accuracy, compared to $0.43$ using the majority baseline, suggesting that 
	the claims are to some extent indicative of the type, but a strong system would need to 
	look at the evidence as well.

	\section{Baseline Model}
	\label{sec:baseline}
	%Our baseline model consists of a retriever module and a verdict prediction module. The 
	%retriever module 
	%retrieves the top $k$ sentences and tables. For each table, relevant cells are selected by 
	%linearizing the table and treating the cell selection as a sequence labelling task. The 
	%retrieved evidence pieces and their contexts are then used as input to a pre-trained 
	%RoBERTa \citep{liu2019roberta} model to predict the correct label using cross-attention 
	%between the different evidence types. %The approach is depicted in Figure 
	%%\ref{fig:baseline}.
	
	\begin{comment}
	\begin{figure}[h]
	\centering
	\includegraphics[width=\textwidth]{images/baseline_new.pdf}
	\caption{Visualisation of the FEVEROUS baseline.}
	\label{fig:baseline}
	\end{figure}
	\end{comment}

	\paragraph{Retriever}
	
	Our baseline retriever module is a combination of entity matching and TF-IDF using DrQA 
	\citep{chen2017reading}. Combining both has previously been shown to work well, 
	particularly for retrieving tables \citep{schlichtkrull2020joint}. We first extract the top 
	$k$ pages by matching extracted entities from the claim with Wikipedia articles. If less 
	than $k$ pages have been identified this way, the remaining pages are selected by Tf-IDF 
	matching between the introductory sentence of an article and the claim. The top $l$ 
	sentences and $q$ tables of the selected pages are then scored separately using TF-IDF. We 
	set $k=5$, $l=5$ and $q=3$. %Retrieving documents first has been important as for tables
	%Combining entity matching with TF-IDF scores led to better results in our preliminary 
	%experiments than a TF-IDF retriever in isolation, as used in FEVER.
	
	For each of the $q$ retrieved tables, we retrieve the most relevant cells by linearizing 
	the table and treating the retrieval of cells as a binary sequence labelling task. The 
	underlying model is 
	a fine-tuned RoBERTa model with the claim concatenated with the respective table as input.  
	When fine-tuning, we deploy row sampling, similar to \citet{oguz2020unified}, to ensure 
	that the tables used during training fit into the input of the model. % We additionally 
	%sample negative tables by using tables from the page of the evidence table.
	
	\paragraph{Verdict prediction}
	Given the retrieved evidence, we predict the verdict label using a RoBERTa encoder with a 
	linear layer on top. Table cells are linearized to be used as a evidence,
	%This decision is based on observations by
	following \citet{schlichtkrull2020joint} who showed that a RoBERTa based model with the 
	right linearization performs better than models taking table structure into account. 
	%Table-BERT~\citep{chenTabFactLargescaleDataset2020} as well as logic-based querying, e.g. 
	%\citet{zhong-etal-2020-logicalfactchecker}. % Table-Bert 
	%%\citep{chenTabFactLargescaleDataset2020}.
	Linearization of a table's content enables cross-attention between cells and sentences by 
	simply concatenating all evidence in the input of the model. For each piece of evidence, we 
	concatenate its context ensuring that the page title appears only once, at the beginning of 
	the evidence retrieved from it. %excluding the title, due to the title already being part 
	%of the context of each element 
	% and thus would result redundancy in the input
	%. Instead, the evidence is grouped per page and separated using a unique token. Each page 
	%group of evidence starts with the page name. 
	
	The verdict predictor is trained on labelled claims with associated cell and sentence 
	evidence and their context.
	%\footnote{For fine-tuning all models we only used a subset of the training data ($20,000$ 
	%samples) as not all annotations were generated by the time the baseline was built.}.
	The FEVEROUS dataset
	is rather %has a very high label
	imbalanced regarding NEI labels (5\% of claims), so we sample additional NEI instances for 
	training
	%. NEI instances are sampled 
	by modifying annotations that contain both cell and sentence evidence by removing either a 
	sentence or an entire table. %Removing an entire table prevents a verdict to become 
	%refuted, e.g.\ when the number of evidence cells is used as evidence.
	We additionally explore the use of a RoBERTa model that has been pre-trained on various NLI 
	datasets (SNLI \citep{bowman-etal-2015-large}, MNLI \citep{williams-etal-2018-broad}, and 
	an NLI-version of FEVER, proposed by \citet{nie-etal-2020-adversarial}).
	
	%The RoBERTa model is fine-tuned using categorical cross-entropy loss.
	
	\section{Experiments}
	
	\subsection{Dataset splits and evaluation}
	\label{sec:evaluation}
	The dataset is split into a training, development and test split in a ratio of about $0.8, 
	0.1, 0.1$. We further ensured that all three claims generated from a highlight are assigned 
	to the same split to prevent claims in the development and test splits from being too 
	similar to the ones in training data.
	%, an issue known to occur in  multiple question answering datasets 
	%\citep{lewis-etal-2021-question}. 
	Quantitative characteristics are shown in Table \ref{tab:split-chracteristics}. Due to the 
	scarcity of NEI instances, we maintained
	%an approximate 
	a rough 
	label balance only for the test set. 
	%This contrasts FEVER, where the number of NEI instances is roughly balanced in each split. 
	%%% to the adjusted definition of NEI, as described above
	In all splits, the number of evidence sets with only sentences as evidence is slightly 
	higher than sets that contain only cell evidence or sets that require a combination of 
	different evidence types. %For both development and test split we further ensured that all 
	%instances require not more than $5$ sentence annotations and $25$ cell annotations limit 
	%the number of pieces of evidence required to verify a
	
	\begin{table}[ht!]
		\centering
		\caption{Quantitative characteristics of each split in FEVEROUS. \\
		}\label{tab:split-chracteristics}
		
		\begin{tabular}{l| ccc | c}
			\toprule
			& \textbf{Train} & \textbf{Dev} & \textbf{Test} & \textbf{Total}  \\
			\midrule
			Supported & 41,835 (59\%)  & 3,908 (50\%)  & 3,372 (43\%) & 49,115 (56\%) \\
			Refuted & 27,215 (38\%) & 3,481 (44\%)  &  2,973 (38\%) & 33,669 (39\%)\\ 
			NEI & 2,241 (3\%) &  501 (6\%) & 1,500 (19\%) & 4,242 (5\%) \\ 
			\midrule 
			Total & 71,291 & 7,890 &  7,845 & 87,026  \\  
			\midrule
			E$_{Sentences}$ & 31,607 (41\%) & 3,745 (43\%)  & 3589 (42\%) & 38,941 (41\%)\\ % 
			%77492 8681   (8467)
			E$_{Cells}$ & 25,020 (32\%) & 2,738 (32\%) &  2816 (33\%) & 30,574 (32\%)\\ 
			E$_{Sentence+Cells}$ & 20,865 (27\%) & 2,468 (25\%)  & 2062 (24\%) & 25,395 
			(27\%)\\ 		
			\bottomrule
		\end{tabular}
		
	\end{table}
	
	%\subsection{Evaluation}
	
	The evaluation considers the correct prediction of the verdict as well as the correct 
	retrieval of evidence. Retrieving relevant evidence is an important requirement, given that 
	it provides a basic justification for the label, which is essential to convince the users 
	of the capabilities of a verification system and to assess its correctness 
	\citep{Uscinski2013TheEO,lipton2016mythos, Kotonya2020ExplainableAF}. Without evidence, the 
	ability to detect machine-generated misinformation is inherently limited 
	\citep{Schuster2020TheLO}. %Thus, in contrast to FEVER where evidence was not evaluated for 
	%instances labelled as NEI, in FEVEROUS we evaluate labels and evidence for all classes. 
	The FEVEROUS score is therefore defined for an instance as follows: 
	\begin{align}
	Score(y,\hat{y}, \mathbb{E}, \hat{E}) = \begin{cases} 1 & \exists{E\in\mathbb{E}} :  E 
	\subseteq \hat{E}  \wedge \text{$\hat{y} = y$},\\
	0 &\text{otherwise}
	\end{cases}
	\label{eq:1}
	\end{align}
	with $\hat{y}$ and $\hat{E}$ being the predicted label and evidence, respectively, and 
	$\mathbb{E}$ the collection of gold evidence sets. Thus, a prediction is scored $1$ iff at 
	least one complete evidence set $E$ is a subset of $\hat{E}$ and the predicted label is 
	correct, % $\exists E \in \mathbb{E}. e\in E  \subset \hat{E}$,
	else 0.
	The rationale behind not including precision in the score is that we recognise that the 
	evidence annotations are unlikely to be exhaustive, and measuring precision would thus 
	penalize potentially correct evidence that was not annotated. Instead, we set an upper 
	bound on the number of elements to be allowed in $\hat{E}$ to
	%. %This evaluation has already been employed at the FEVER shared task in 2018 and was 
	%%received well. 
	%The restrictions for $\hat{E}$ is set to 
	$s$ table cells and $l$ sentences. %of cell evidence and and other types of evidence, 
	%respectively.
	This distinction was made because the number of cells used as evidence is typically higher 
	than the number of sentences. $s$ is set to $25$ and $l$ to $5$, ensuring that the upper 
	bound covers the required number of evidence pieces for every instance $E$ in both 
	development and test set. 
	
	%%\paragraph{Shared task plan}
	The FEVEROUS dataset was used for the shared task of the FEVER Workshop 2021 
	\citep{aly-etal-2021-fever}, with the same splits and evaluation as presented in this 
	paper. %The primary evaluation metric is the FEVEROUS score (Eq. \ref{eq:1}), with evidence 
	%$F_1$ being considered in the case of ties between systems. 
	\begin{comment}
	To ensure that the potential to fit the test set is minimal, the test set will remain 
	hidden until the last week of the shared task, when only the claims will be released to the 
	participants to run their systems on and submit their answers. After the end of the shared 
	task, the evaluation will remain open to encourage further work on FEVEROUS, however the 
	test set annotation shall remain hidden.
	\end{comment}
	
	\subsection{Results}
	Table \ref{tab:final-scores} shows the results of our full baseline compared to a 
	sentence-only and a table-only baseline.%\footnote{For fine-tuning all models we only used 
	%a subset of the training data ($20,000$ samples) as not all annotations were available 
	%when 
	%they experiments were run.} 
	% Both sentence-only and table-only baseline use the TF-IDF retriever (c.f. section 
	%\ref{sec:baseline}). While the sentence-only baseline and verdict predictor)
	All baselines use our TF-IDF retriever with the sentence-only and table-only baseline 
	extracting sentences and tables only, respectively. While the sentence-only model predicts 
	the verdict label using only extracted sentences, the %to predict the label using a 
	%multilayer perceptron,
	the table-only baseline only extracts the cells from retrieved tables with our cell 
	extractor model and predicts the verdict by linearising the selected cells and their 
	context. All models use our verdict predictor for classification. %these cells and using 
	%them with our RoBERTa classifier. 
	Our baseline that combines both tables and sentences achieves substantially higher sores 
	than when focusing exclusively on either sentences or tables. %Interestingly,  % performs 
	%%better  combining both sentences and tables % linearizes them our RoBERTa classifier to 
	%%%predict the verdict.

	\begin{table}[ht!]
		\centering
		\caption{FEVEROUS scores for the sentence-only, table-only, and full baseline for both 
		development and test set. \emph{Evidence} measures the full coverage of evidence (i.e. 
		Eq. \ref{eq:1} without the condition on correct prediction $\hat{y}=y$).\\
		}
		\label{tab:final-scores}
		\begin{tabular}{c|cc|cc}
			\toprule
			\multirow{2}{*}{\textbf{Model}} & \multicolumn{2}{c|}{\textbf{Dev}} & 
			\multicolumn{2}{c}{\textbf{Test}}  \\
			& Score & Evidence & Score & Evidence \\
			\toprule
			Sentence-only baseline & 0.13  & 0.19 & 0.12 & 0.19  \\ %0.14 F1 dev 0.16 Test
			%Sentence-only baseline & 0.08  & 0.14 & 0.09 & 0.19
			%Table-only baseline & 0.03 & 0.08 & 0.04 & 0.06 \\
			%Table-only baseline & 0.04 & 0.05 & 0.04 & 0.06 \\ USING FIRST 25 cells
			Table-only baseline & 0.05 & 0.07 & 0.05 & 0.07 \\
			\midrule
			Full baseline & 0.19 & 0.29 & 0.18 & 0.29 \\
			\bottomrule
		\end{tabular}
	\end{table}
	
	\paragraph{Evidence Retrieval}
	To measure the performance of the evidence retrieval module for retrieved evidence 
	$\hat{E}$, we measure both the Recall@k %$\frac{|E \cap \hat{E}|}{|E|}$
	on a document level as well as on a passage level (i.e.\ sentences and tables).  %The 
	%document coverage of DPR is the set of articles selected passages belong to.
	Results are shown in Table \ref{tab:sub-results}.
	As seen for $k=5$ the retriever achieves a document coverage of $69\%$. The top $5$ 
	retrieved sentences cover $53\%$ of all sentences while the top $3$ tables have a coverage 
	of $56\%$, highlighting the effectiveness of our retriever to retrieve both sentences and 
	tables. The overall passage recall is $0.55\%$. For comparison a TF-IDF retriever without 
	entity matching achieves a coverage of only $49\%$.
	
	Extracting evidence cells when the cell extraction model is given the gold table for the 
	claim from the annotated data leads % tables that contain cell evidence $e \in E$
	%for a given claim leads
	to a cell recall of $0.69$, with a recall of $0.74$ when a table contains only a single 
	cell as evidence. %For comparison using simply the first $25$ cells of the gold table 
	%achieves a recall of $X$. 
	Extracted cells from the retrieved tables in combination with the extracted sentences fully 
	cover the evidence of 29\% samples in the dev set.

	\paragraph{Verdict prediction}

	The right side of Table \ref{tab:sub-results} shows oracle results (i.e.\ when given the 
	correct evidence), as well as results without NEI sampling and without an NLI pre-trained 
	model. Without the NEI sampling, the model is not able to recognise a single NEI sample 
	correctly. NLI pre-training further increases results, resulting in a macro-averaged $F_1$ 
	of $0.70$. 
	
	\begin{table}[ht!]
		\centering
		\caption{(left) Document and passage (sentence + tables) coverage for the retrieval 
		module. (right) %Oracle
			Verdict classification using gold evidence. \emph{NLI} denotes pre-training on NLI 
			corpora and \emph{NEI} NEI sampling.  Scores are reported in per-class $F_1$. The 
			overall score is reported using macro-averaged $F_1$. All results are reported on 
			the dev set. \\
		}\label{tab:sub-results}
		\resizebox{1\linewidth}{!}{
			\begin{tabular}{ cc }
				\begin{tabular}{l | ccc}
					\toprule
					\textbf{top}   & \textbf{Doc (\%)} & \textbf{Sent (\%)} &  \textbf{Tab 
					(\%)}  \\
					% k  & Document Coverage (\%) & Passage coverage (\%)  \\
					\toprule
					1  &  0.39 & 0.23 & 0.45   \\
					2  & 0.49 &   0.37   & 0.54 \\ 
					3  & 0.58 &   0.46  & 0.56  \\ 
					5  &  0.69 &   0.53 & -  \\ 
					\bottomrule
				\end{tabular} 
				&
				\begin{tabular}{l| cccc}
					\toprule
					\textbf{Model}  & \textbf{Supported} & \textbf{Refuted} & \textbf{NEI} & 
					\textbf{Overall}  \\
					\toprule
					RoBERTa  & 0.89 & 0.87  & 0.05 & 0.53  \\ 
					\hspace{0.4em}+NLI &  0.90 & 0.88 & 0.09 & 0.62\\
					\hspace{0.4em}+NLI+NEI  & 0.89 & 0.87 & 0.34 & 0.70\\ 
					
					\bottomrule
				\end{tabular} \\
		\end{tabular}}
		
	\end{table}
	
	\subsection{Discussion}
	\label{sec:discussion}

	\textbf{Retrieval of structured information.} % To further assess to which extent our 
	While the verdict predictor combines information from both evidence types, our retrieval 
	system extracts structured and unstructured information largely independently. However, 
	tables are often 
	%serve a supporting purpose, being 
	specified and described by surrounding sentences. For instance 
	\citep{zayats-etal-2021-representations} enhance Wikipedia table representations by using 
	additional context from surrounding text. Thus, sentences provide important context to 
	tables to be understood and related to the claim (and vice versa). Moreover, we have 
	ignored hyperlinks in our model, yet they are excellent for entity grounding and 
	disambiguation, adding context to both tables and sentences. % Taking into account  and 
	%additional information of unstructued information and
	
	\textbf{Numerical Reasoning.} An aspect currently ignored by our baseline is that a 
	substantial number of claims in FEVEROUS require \emph{numerical reasoning} (for about 10\% 
	of claims numerical reasoning was selected as the main verification challenge), ranging 
	from simply counting matching cells to arithmetic operations. \citet{dua-etal-2019-drop} 
	showed that reading comprehension models lack the ability to do simple arithmetic 
	operations. 
	%Covering these cases in FEVEROUS is left for future work. 

	\textbf{Verification of complex claims.} Compared to previous datasets, the length of 
	claims and number of required evidence is substantially higher. As a result, more pieces of 
	evidence per claim need to be retrieved
	%information in the retrieval corpus needs to be considered 
	and related to each other. 
	%Moreover, verifying longer claims requires to understand which information from a claim 
	%needs be fact-checked. %How should a claim be broken down into its components that have to 
	%%be fact-checked?
	This opens opportunities to explore %also raises the question on the importance of 
	the effect of the order in which each part of a claim is being verified and how evidence is 
	conditioned on each other. To facilitate research in this direction, FEVEROUS contains for 
	each annotation a list of operations (e.g.\ searched ..., clicked on hyperlink ...) that an 
	annotator used to verify a claim (see supplementary material). % Since the FEVEROUS 
	
	\textbf{Ecological Validity} Incorporating information from both text and tables for 
	fact-checking enables the verification of more complex claims than previous large-scale 
	datasets, ultimately enhancing the practical relevance of automated fact-checking systems. 
	However, FEVEROUS still simplifies real-world claims substantially, by controlling many 
	variables of the claim generation process. For instance, it ignores the common strategy of 
	biased evidence employed for generating misleading claims in the real world, also referred 
	to as cherry picking, where facts which are true in isolation are being taken out of 
	context, resulting in an overall false claim. 
	
	\section{Conclusion}
	This paper introduced FEVEROUS, the first large-scale dataset and benchmark for fact 
	verification that includes both unstructured and structured information. We described the 
	annotation process and the steps taken to minimise biases and dataset artefacts during 
	construction, and discussed aspects in which FEVEROUS differs from other fact-checking 
	datasets. We proposed a baseline that retrieves sentences and table cells to predict the 
	verdict using both types of evidence, and showed that it outperforms sentence-only and  
	table-only baselines. With the baseline achieving a score of $0.18$ %a FEVEROUS score of 
	%$0.10$
	we believe that FEVEROUS is a challenging yet attractive benchmark for the development of 
	fact-checking systems.% as well as other directions as indicated in our discussion.% many 
	%%other directions to be and many other aspects to be discovered. systems.
	
	\section*{Acknowledgements}
	
	We would like to thank Amazon for sponsoring the dataset generation and supporting the 
	FEVER workshop and shared task. The dataset generation and its public release was 
	coordinated by the University of Cambridge.
	Rami Aly is supported by the Engineering and Physical Sciences Research Council Doctoral 
	Training Partnership (EPSRC).
	James Thorne is supported by an Amazon Alexa Graduate Research Fellowship.
	Zhijiang Guo, Michael Schlichtkrull and Andreas Vlachos are supported by the ERC grant 
	AVeriTeC (GA 865958).

	\bibliography{bibliography.bib}

\section{Supplementary Material}
\subsection{Access to the dataset}
The FEVEROUS dataset can be accessed from the official website of the FEVER Workshop 
\url{https://fever.ai/dataset/feverous.html} and is hosted in an the same AWS S3 Bucket as the 
FEVER dataset, which has been publicly available since 2018. As the authors of this paper 
manage the workshop's website they can ensure proper maintenance and access to the dataset. The 
hosting of the dataset includes a retrieval corpus, as well as each split of the dataset. At 
the time of this paper's submission, the shared task is still ongoing with the unlabeled test 
set being kept hidden until the last week of the shared task. %This includes reviewers so that 
%they are eligible to participate in the shared task, if interested.
%Code of both baseline and annotation platform can be accessed through.
Elementary code to process the data from both annotations and the provided Wikipedia DB (e.g. 
extracting context for a given element, getting a table from a cell ID etc..) is publicly 
available on \url{https://github.com/Raldir/FEVEROUS}. The repository also contains the code of 
the annotation platform as well as the baseline's code. The DOI for the FEVEROUS dataset is 
\texttt{10.5281/zenodo.4911508} and structured metadata has been added to the webpage.% While 
%not yet available, we will add structured metadata for FEVEROUS very soon as well.  % and have 
%%just started the process for it.  %with the aim to make the dataset more accessible as we 
%%%recognize that its format and content are comparably complex.

The training and development data is hosted in Jsonlines format. Jsonlines contains a single 
JSON per line, encoded in UTF-8. This format allows to process one record at a time, and works 
well with unix/shell pipelines. Each entry consists of five fields:
The training and development data contains 5 fields:
\begin{itemize}
	\item id: The ID of the sample
	\item label: The annotated label for the claim. Can be one of SUPPORTS|REFUTES|NOT ENOUGH 
	INFO.
	\item claim: The text of the claim.
	\item evidence: A list (at maximum three) of evidence sets. Each set consists of 
	dictionaries with two fields (content, context).
	\begin{itemize}
		\item content: A list of element ids serving as the evidence for the claim. Each 
		element id is in the format "[PAGE ID]\_[EVIDENCE TYPE]\_[NUMBER ID]". [EVIDENCE TYPE] 
		can be sentence, cell, header\_cell, table\_caption, item.
		\item context: A dictionary that maps each element id in content to a set of Wikipedia 
		elements that are automatically associated with that element id and serve as context. 
		This includes an article's title, relevant sections (the section and sub-section(s) the 
		element is located in), and for cells the closest row and column header (multiple 
		row/column headers if they follow each other).
	\end{itemize}
	\item annotator\_operations: A list of operations an annotator used to find the evidence 
	and reach a verdict, given the claim. Each element in the list is a dictionary with the 
	fields (operation, value, time).
	\begin{itemize}
		\item operation: Any of the following
		\begin{itemize}
			\item start, finish: Annotation started/finished. The value is the name of the 
			operation.
			\item search: Annotator used the Wikipedia search function. The value is the 
			entered search term or the term selected from the automatic suggestions. If the 
			annotator did not select any of the suggestions but instead went into advanced 
			search, the term is prefixed with "contains..."
			\item hyperlink: Annotator clicked on a hyperlink in the page. The value is the 
			anchor text of the hyperlink.
			\item Now on: The page the annotator has landed after a search or a hyperlink 
			click. The value is the PAGE ID.
			\item Page search: Annotator search on a page. The value is the search term.
			\item page-search-reset: Annotator cleared the search box. The value is the name of 
			the operation.
			Highlighting, Highlighting deleted: Annotator selected/unselected an element on the 
			page. The value is ELEMENT ID.
			\item back-button-clicked: Annotator pressed the back button. The value is the name 
			of the operation.
			\item value: The value associated with the operation.
			\item time: The time in seconds from the start of the annotation.
		\end{itemize}
	\end{itemize}
	\item expected\_challenge: The challenge the claim generator selected will be faced when 
	verifying the claim, one out of the following: Numerical Reasoning, Multi-hop Reasoning, 
	Entity Disambiguation, Combining Tables and Text, Search terms not in claim, and Other.
	\item challenge: The main challenge to verify the claim, one out of the following: 
	Numerical Reasoning, Multi-hop Reasoning, Entity Disambiguation, Combining Tables and Text, 
	Search terms not in claim, and Other.
\end{itemize}

The retrieval corpus is provided to annotators in either Jsonlines format, or as an SQLite3 
database. The latter allows faster retrieval for articles by their name, which is helpful for 
instance when mapping annotation ids to their contents.  Each Wikipedia article contains 2 base 
fields:

\begin{itemize}
	\item title: The title of the Wikipedia article
	\item order: A list of elements on the Wikipedia article in order of their appearance. 
	Elements can be: section, table, list, sentence.
\end{itemize}

Each element specified in order is a field. A sentence field contains the text of the sentence.

A section element is a dictionary with following fields:

\begin{itemize}
	\item value: Section text
	\item level: The level/depth of the section.
\end{itemize}

A table element is a dictionary with following fields:
\begin{itemize}
	\item type: Whether the table is an infobox or a normal table
	\item table: The content of the table. The table is specified as a list of lists. Each 
	element in a list is a cell with the fields (id, value, is\_header, row\_span, 
	column\_span).
	\item caption: Only specified if the table contains a caption.
\end{itemize}

A list element consists of following fields:

\begin{itemize}
	\item type: Whether the list is an ordered or unordered list
	\item list: A list of dictionaries, with fields being (id, value, level, type). level is 
	the depth of the list item. The level increments with each nested list. type specifies type 
	of a nested list, which is starting after the item specifying the type. Field is only 
	specified if the next item is in a nested list.
	Hyperlinks in text are indicated with double square brackets. If an anchor text is 
	provided, it is the text on the right hand side of a vertical bar in the square backets
\end{itemize}

Example dataset and retrieval corpus entries can be found on 
\url{https://fever.ai/dataset/feverous.html}.

\subsection{Ethics statement}

The FEVEROUS dataset was collected with approval and following the practices outlined by the 
Ethics Committee of the Computer Lab of the University of Cambridge (reference number 1842).  
Furthermore, the external contractor has a well-outlined policy regarding their code of ethics 
to ensure the well-being of all annotators in our experiment. Their Code of Ethics consists of: 
Fair Pay, Inclusion, Crowd Voice (i.e. Feedback mechanisms), Privacy and Confidentiality, 
Communication, and Well-Being. 

We anticipate that FEVEROUS will be used for the development of fact checking systems that 
might be applied in real world contexts to assign truth/false labels, similar to those on fact 
checking websites run by journalists. We use the labels supported/refuted (by evidence) instead 
of true/false to be clear that we do not make any judgements about the truth of a statement in 
the real-world, but only consider Wikipedia as the source of evidence to be used. And while 
Wikipedia is a great collaborative resource, it has mistakes and noise of its own similar to 
any encyclopedia or knowledge source. Thus we discourage users of FEVEROUS to make absolute 
statements about the claims being verified, i.e.\ avoid using it to develop truth-tellers. 
Finally, we require systems to predict when the evidence is not sufficient to make a judgement, 
in which case it would be useful to look beyond Wikipedia for evidence.

We did not collect personal data of the participants in any way. A participant is only 
identified using an identification number to access our online tool. Generated claims must only 
include information on Wikipedia or considered to be general world knowledge, while all 
evidence is taken from Wikipedia directly, thus not including any personally identifiable 
information or offensive content.

%: \url{https://appen.com/crowd-wellness/}.

\subsection{Data Statements}
We follow the data statements structure of \citet{bender_data_statements2018} to give 
additional insights into the dataset and its construction.
\paragraph{Curation Rationale.} In order to study fact extraction and verification on both 
unstructured and structured information, we use the entire English Wikipedia as the knowledge 
base. Wikipedia is a large-scale collaboratively created encyclopedia, covering a large extent 
of knowledge/topics and is as such considered to be a suitable testbed for our purpose. Only 
articles that have been flagged by Wikipedia to have issues, miss references and/or citations 
have been excluded. The rationale behind this decision is to compile a retrieval corpus with 
information that is consistent across pages. The entire content of an article is considered, 
with exception to sections that were flagged as aforementioned as well sections that are named 
'References', 'Citations', 'Sources', 'Further reading', 'External links', 'Works', 'Gallery', 
'Citations and references', 'Bibliography', or 'External links \& References' as we consider 
these sections to be out-of-scope for our task. Sentence and Table highlights given to 
annotators were sampled randomly from the entire collection of English Wikipedia articles.

\paragraph{Language Variety.} The extracted evidence aligns with English Wikipedia's 
characteristic on language variety. A section on this, describing the lack of standardization 
can be found \href{https://en.wikipedia.org/wiki/English_Wikipedia}{here}. For claim 
generation, half of the annotators were native US-English speakers, while the other half were 
English speakers from the Philippines. For claim verification, all annotators were native 
US-English speakers. The internal screening by the external contractor ensured that the variety 
of English used is very similar across annotators, being en-us.

\paragraph{Speech Situation.} The retrieval corpus was compiled based on a December 2020 
version of English Wikipedia. Wikipedia is a collaborative encyclopedia, and as such regularly 
edited. Wikipedia describes in detail the requirements and recommendation of texts in articles, 
which can be found for instance 
\href{https://en.wikipedia.org/wiki/Help:Your_first_article}{here}, 
\href{https://en.wikipedia.org/wiki/Wikipedia:Writing_better_articles}{here}, and 
\href{https://en.wikipedia.org/wiki/Wikipedia:Policies_and_guidelines}{here}. Claims were 
generated between March and May 2021, with very detailed guidelines regarding content and 
structure. A claim is described in the guidelines as \emph{ a single well-formed sentence. It 
should end with a period; it should follow correct capitalization of entity names (e.g. 
`India', not `india'); numbers can be formatted in any appropriate English format (including as 
words for smaller quantities).} They further \emph{must \textbf{\underline{not be subjective}} 
and be \textbf{\underline{verifiable}} using publicly available information/knowledge}. Claims 
further should as unambiguous as possible, and must not contain any idioms, figures of speech, 
similes, or verbose language (see Section \ref{sec:claim-generation}). 

\paragraph{Text characteristics.} Since highlights were sampled randomly from Wikipedia 
articles, the distribution of topics of generated claims roughly corresponds to the underlying 
English Wikipedia distribution of articles (i.e. people, geography, history, and sports being 
the main topics). We restrict the topic in some instances, such as: \emph{Claims should not be 
about contemporary political topics  (e.g. contemporary Wars (from the second world war and 
onwards), or disputed topics)}.

\paragraph{Annotator Demographic}
Annotator candidates were screened specifically for our task, with multiple screening and 
calibration-stages as described in the paper. This ensures that annotators are aware of the 
constraints and guidelines when generating claims and verifying them.  All annotators were paid 
above their local minimum wage.
\begin{itemize}
	\item Age: Claim generation: 11 people between 18-24 years, 20 people between 25-34 years, 
	8 people between 35-44 years, 6 people between 45-54, and 12 people unspecified. Claim 
	verification: 4 people between 18-24 years, 17 people between 25-34 years, 9 people between 
	35-44 years, 12 people between 45-54, 5 people between 55-64, and 12 people unspecified. 
	\item Gender: Claim generation: 11 male, 42 female, and 4 unspecified. Claim verification: 
	15 male, 36 Female, and 3 unspecified.
	\item Race/ethnicity: -
	\item Native language: Claim generation: 33 people are native en-us speakers, 24 annotators 
	are native en-ph (English (Philippines)) speaker. Claim verification: All annotators are 
	native en-us speakers.  
	\item Socioeconomic status: -
	\item  Training in linguistics/other relevant discipline:  Claim generation: English 
	speakers from the Philippines are language-aware (an upper education degree in a 
	language-related subject). Claim verification: all annotators are language-aware.
\end{itemize}

\subsection{Licensing}
These data annotations incorporate material from Wikipedia, which is licensed pursuant to the 
\href{https://en.wikipedia.org/wiki/Wikipedia:Copyrights}{Wikipedia Copyright Policy}. These 
annotations are made available under the license terms described on the applicable Wikipedia 
article pages, or, where Wikipedia license terms are unavailable, under the Creative Commons 
Attribution-ShareAlike License (version 3.0), available at 
\href{http://creativecommons.org/licenses/by-sa/3.0/}{\url{http://creativecommons.org/licenses/by-sa/3.0/}}
 (collectively, the “License Terms”). You may not use these files except in compliance with the 
applicable License Terms. Credits to the contents of a page go to the authors of the 
corresponding Wikipedia article. Since article names in the dataset are unchanged, the authors 
can be found on the respective article on Wikipedia (\url{https://www.wikipedia.org/wiki} + 
TITLE\_ID). The associated code to FEVEROUS (i.e. annotation platform, baseline code) are 
licensed under Apache 2.0.

\subsection{Detailed dataset and annotation statistics}

\subsubsection{Claim generation}
\begin{figure}[ht!]
	\begin{subfigure}[b]{0.5\textwidth}
		\centering
		\includegraphics[width=\textwidth]{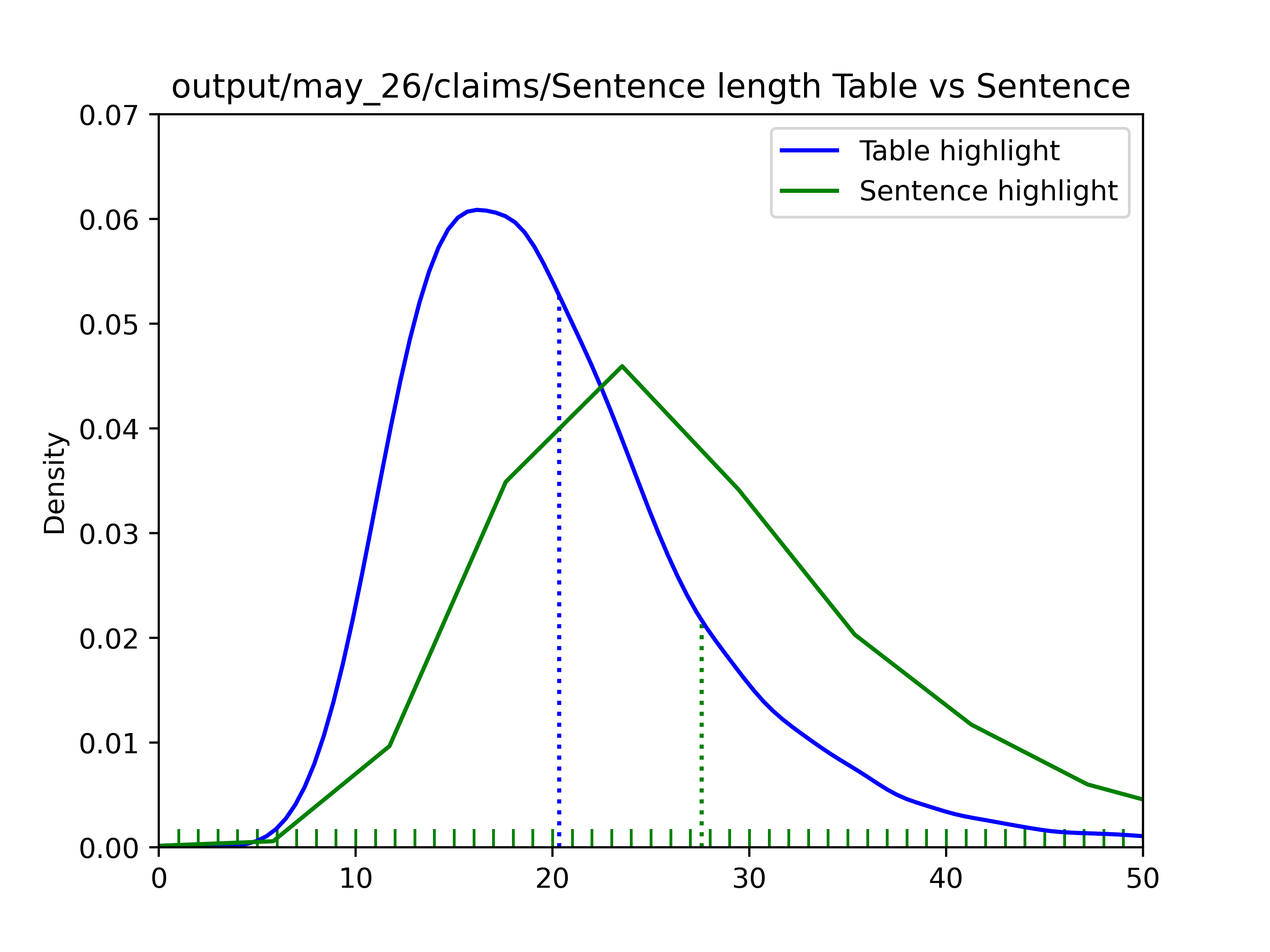}
		\caption{Sentence length of claims given sentence versus table highlight.}
		\label{fig:claim-stats-1}
	\end{subfigure}
	\hfill
	\begin{subfigure}[b]{0.5\textwidth}
		\centering
		\includegraphics[width=\textwidth]{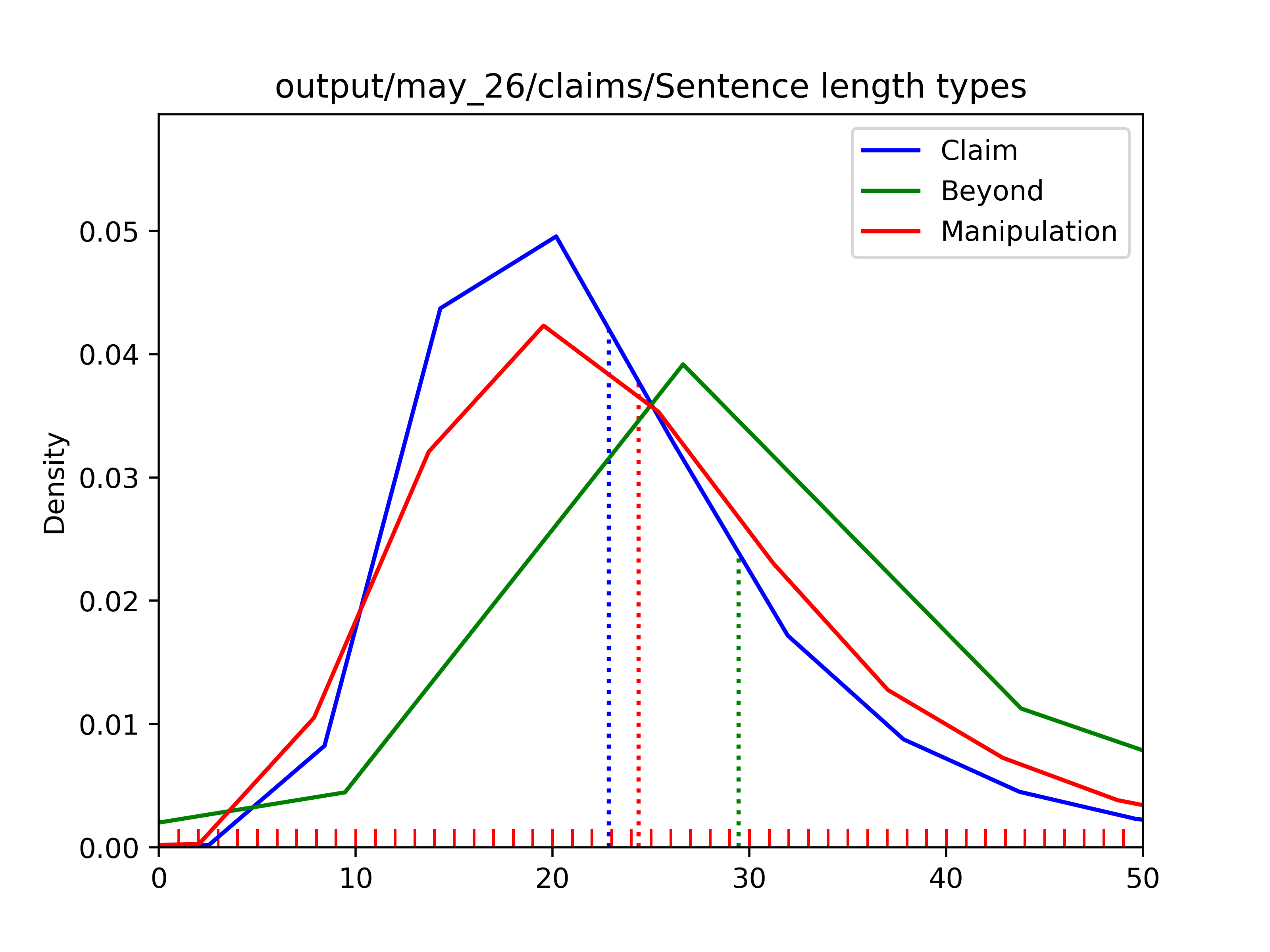}
		\caption{Sentence length for different claim types.}
		\label{fig:claim-stats-2}
	\end{subfigure}
\end{figure}

An average annotation (i.e. generating three claims) took an annotator $373$ seconds. A total 
of $61058$, and $32700$ claims were created using table and sentence highlights, respectively. 
$47300$ annotations were prompted to use information from the same page, and $46428$ from 
different pages. The average length of a claim is $23$, $29$, and $24$ for Type I, Type II and 
Type III, respectively. Annotators used on average $0.71$ hyperlinks and $0.15$ search queries. 
For Type II claims that require multiple pages, annotator used on average $1.2$ hyperlinks and 
$0.2$ searches. Sentence length by claim type is shown in figure \ref{fig:claim-stats-1}. 
Average sentence length for both table and sentence highlights is seen in figure 
\ref{fig:claim-stats-2}.

\subsubsection{Claim verification}
A single claim verification took on average $165$ seconds. Claims are selected uniformly from 
the pool of different claim types, resulting in an claim verification set of about equal claims 
for each claim type. On average an annotation has $1.1$ evidence sets, with a total of $7468$ 
annotations having more than one evidence set. Annotators needed on average $1.34$ search 
queries and $0.72$ hyperlinks. On average $0.1$ advanced searches were used (i.e. searches for 
which no of the given page suggestions matches, so that annotators had to go to the advanced 
search page that uses 'in page' matches with Elasticsearch). In about $84\%$ of claims do the 
pages from which evidence was retrieved directly match a word or phrase in the claim itself. 
$69\%$ of all pieces of evidence are table cells, $29\%$ are sentences, $1\%$ are list items, 
and $1\%$ are table captions. 

\begin{figure}[ht!]
	\begin{subfigure}[b]{0.5\textwidth}
		\centering
		\includegraphics[width=\textwidth]{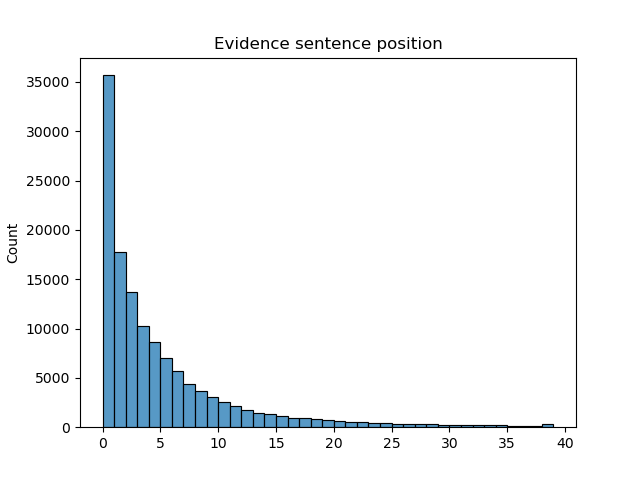}
		\caption{Sentence position distribution in evidence.}
		\label{fig:ev-stats-1}
	\end{subfigure}
	\hfill
	\begin{subfigure}[b]{0.5\textwidth}
		\centering
		\includegraphics[width=\textwidth]{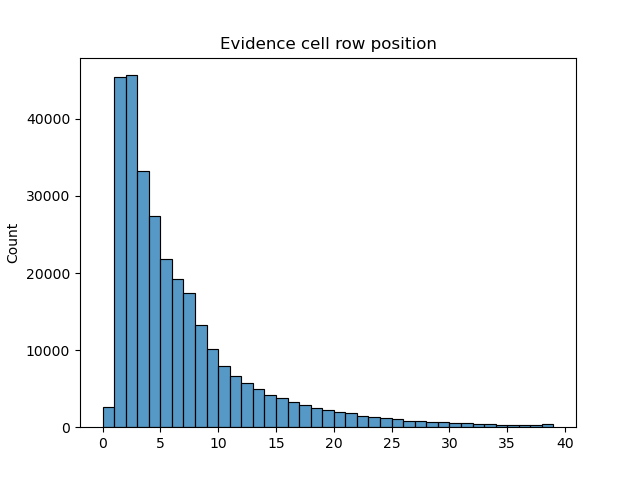}
		\caption{Row position of cells in tables in evidence.}
		\label{fig:ev-stats-2}
	\end{subfigure}
\end{figure}

\begin{figure}[ht!]
	\begin{subfigure}[b]{0.5\textwidth}
		\centering
		\includegraphics[width=\textwidth]{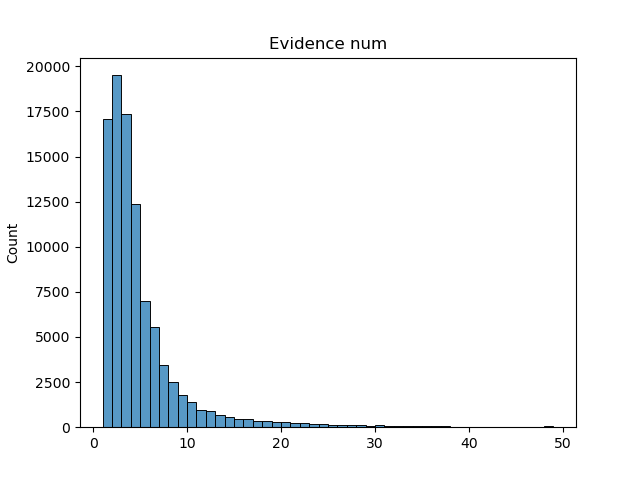}
		\caption{Distribution of number of evidence pieces in an evidence set.}
		\label{fig:ev-stats-3}
	\end{subfigure}
	\hfill
	\begin{subfigure}[b]{0.5\textwidth}
		\centering
		\includegraphics[width=\textwidth]{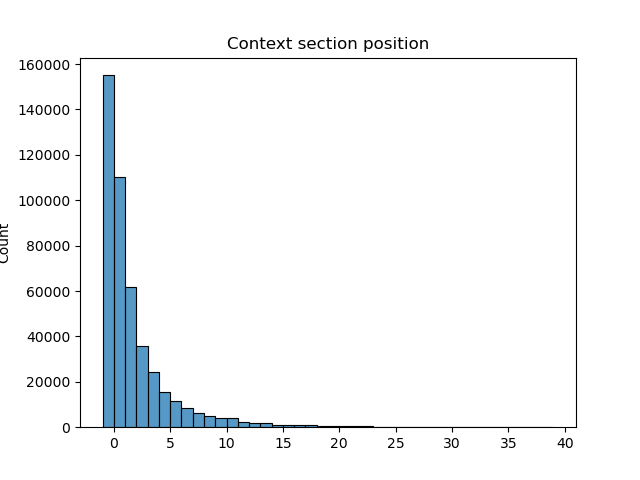}
		\caption{Section position of evidence pieces.}
		\label{fig:ev-stats-4}
	\end{subfigure}
\end{figure}

Plot \ref{fig:ev-stats-1} and \ref{fig:ev-stats-2} show the evidences' sentence positions and 
row positions of cells in tables, respectively. Plot \ref{fig:ev-stats-3} shows the 
distribution of evidence numbers in the dataset. Plot \ref{fig:ev-stats-4} shows the section 
number where evidence is located, with $-1$ being the introduction section.

\subsubsection{Claim Verification Challenges}

\begin{figure}[ht!]
	\centering
	\includegraphics[width=0.8\textwidth]{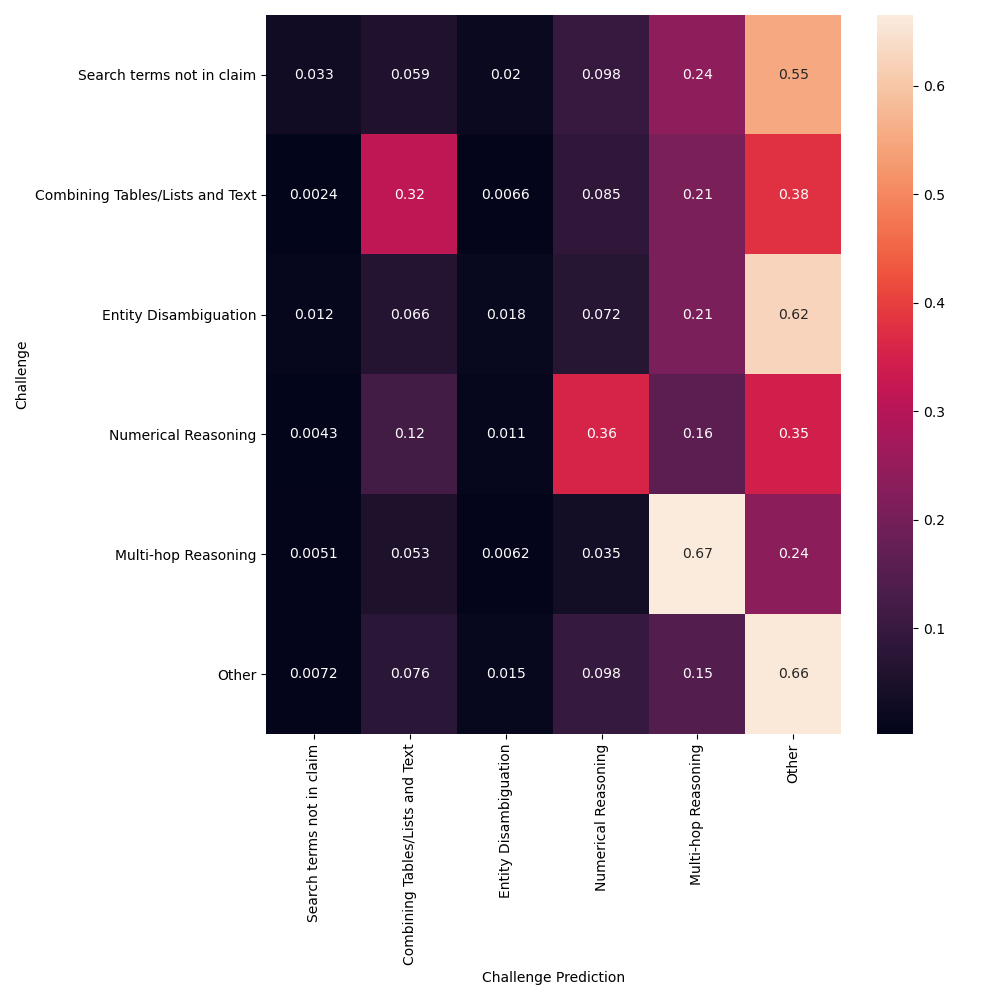}
	\caption{Confusion matrix for expected challenges versus actual challenges. Numbers are 
	normalized across the x-axis.}
	\label{fig:feverous-challenges-conf}
\end{figure}

Table \ref{tab:split-challenges} shows the distribution of verification challenges in the 
FEVEROUS dataset, both the expected challenges as selected by the claim generators as well as 
the verification challenges by the verification annotators. The latter constitutes the actual 
distribution of challenges in FEVEROUS. As seen, the distribution is relatively similar across 
the splits, with about 10\% of all claims having numerical reasoning, 16\% multi-hop 
reasoning,  14\% Combining Tables and text, 2\% Entity Disambiguation, and 1.3\% Search terms 
not in claim as their main verification challenge. Figure \ref{fig:feverous-challenges-conf} 
shows the confusion matrix between expected and actual challenges, normalized along the x-axis. 
It is apparent that the claim generators overpredicted \emph{Other} as the main challenge, 
indicating that the generators were frequently not aware of the challenge their claim poses 
when generating them. This particularly applies to \emph{Entity Disambiguation} and 
\emph{Search terms not in claim}, which have almost never been predicted correctly by the 
generators, most likely due to the generators not searching for the pages themselves, but are 
given a highlight on a page to generate their claim. Interestingly, there is also some 
discrepancy between numerical claims as well as claims that need both tables and text. This 
might be explained by information redundancy, having generated claims using both tables and 
text, not knowing that there is a sentence that contains the information of both.  Further 
analyzing the challenges might lead to highly interesting insights on interactions and 
discrepancies between the expected difficulties from someone generating a claim and an 
annotator actually verifying it.

Following we show an example claim for each challenge category, taken from the dataset:
\begin{itemize}
	\item   \textbf{Numerical Reasoning} \hspace{1em} As of the 2011 Indian census, Nimbapur 
	--located in the Indian state of Maharashtra, which is the second-most populous Indian 
	state -- has a population of 1903, with nearly half of the residents being non-workers. 
	(\textit{Calculation of the ratio between total population and residents who are 
	non-workers})
	\item \textbf{Multi-hop Reasoning} \hspace{1em} Belgium's Léon Schots, a Belgian former 
	long-distance runner who competed in track and cross country running competitions, was the 
	fastest athlete in the senior men's race (12.3km) at the 1977 IAAF World Cross Country 
	Championships. (\textit{Evidence to verify the claim are from two different articles})\\
	\item \textbf{Entity Disambiguation} \hspace{1em} VUKOVI is a rock band from Scotland that 
	plays pop rock, noise pop music and is formerly called Wolves. \textit{Disambiguation of 
	the term Wolves}
	\item \textbf{Search terms not in claim} \hspace{1em} In 2011, Evans signed with the 
	Cincinnati Bengals after going undrafted in the NFL draft; but in November 2011, Evans was 
	suspended for four games. (\textit{To retrieve evidence, annotator first searched for any 
	page containing "Evans signed with the Cincinnati Bengals", until finding the page for the 
	entity's full name "DeQuin Evans".})
	\item  \textbf{Combining Tables and Text} \hspace{1em} Braeden Lemasters, an American 
	actor, musician, and voice actor, appeared in six films since 2008 and also appeared in TV 
	shows such as Six Feet Under where he starred as Frankie. (\textit{Needing evidence from 
	both tables and text})
	\item    \textbf{Other} \hspace{1em} Aquarion Logos is an anime series produced by 
	Satelight which is a Japanese animation studio which serves as a division of pachinko 
	operator Symphogear  Group. (\textit{Neither of the above five challenges apply})
\end{itemize}

\begin{table}[ht!]
	\centering
	%	\resizebox{\columnwidth}{!}{%
	\begin{tabular}{l| cccc }
		\toprule
		Challenge Category  & \textbf{Train} & \textbf{Dev} & \textbf{Test} & \textbf{Total}\\
		\midrule
		\multicolumn{5}{c}{\textbf{Expected Challenges}}\\
		\midrule
		Numerical Reasoning & 7798  & 1024 & 842 & 9664\\
		Multi-hop Reasoning & 17248 & 1871  &  2011 & 21130\\ 
		Entity Disambiguation & 826 &  143& 77 & 1046  \\ 
		Combining Tables and Text & 7775 & 975 &  769 & 9519 \\
		Search terms not in claim & 405 & 57 & 90 & 552 \\
		Other & 37239 & 3820 & 4056 & 45115\\
		\midrule
		\multicolumn{5}{c}{\textbf{Verification Challenges}}\\
		\midrule
		Numerical Reasoning & 7214  & 873 & 740 & 8827\\
		Multi-hop Reasoning & 11624 & 1281  &  1195 & 14100\\ 
		Entity Disambiguation & 1353 &  201& 200 & 1754  \\ 
		Combining Tables and Text & 10083 & 1035 &  940 & 12,058 \\
		Search terms not in claim & 824 & 131 & 193 & 1148 \\
		Other & 40193 & 4369 & 4577 & 49139 \\
		\bottomrule
	\end{tabular}
	\caption{Distribution of verification challenges in the FEVEROUS dataset. Top: Expected 
	verification challenges, selected during claim generation. Bottom: Verification challenges, 
	selected by annotator after a claim was verified.\\
	}\label{tab:split-challenges}
\end{table}

\subsubsection{Details on QA statistics} 
\begin{figure}[ht!]
	\begin{subfigure}[b]{0.5\textwidth}
		\centering
		\includegraphics[width=\textwidth]{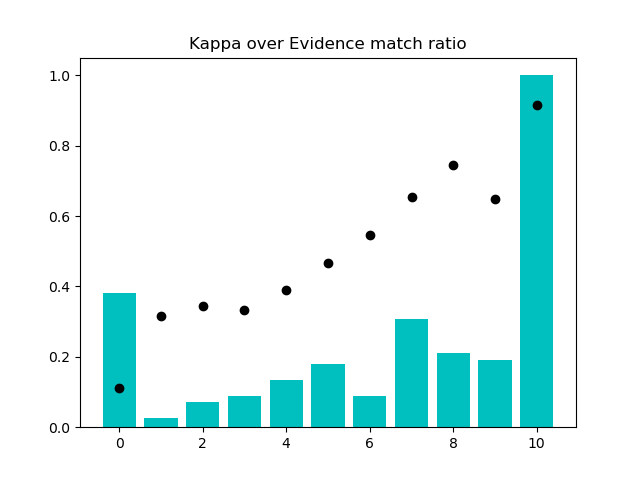}
		\caption{Kappa score $\kappa$ measured for QA-claims with different evidence match 
		ratio. Blue bars indicate the proportion of total annotations with that evidence match. 
		The match is calculated as as $\frac{2* |E_1 \cap E_2|}{|E_1| + |E_2|}$. Shown x-values 
		(similarity) are multiplied by ten.} %as $ \frac{2* |E_1 \cap E_2|}{|E_1| + |E_2|}$}
		\label{fig:qa-stats-1}
	\end{subfigure}
	\hfill
	\begin{subfigure}[b]{0.5\textwidth}
		\centering
		\includegraphics[width=\textwidth]{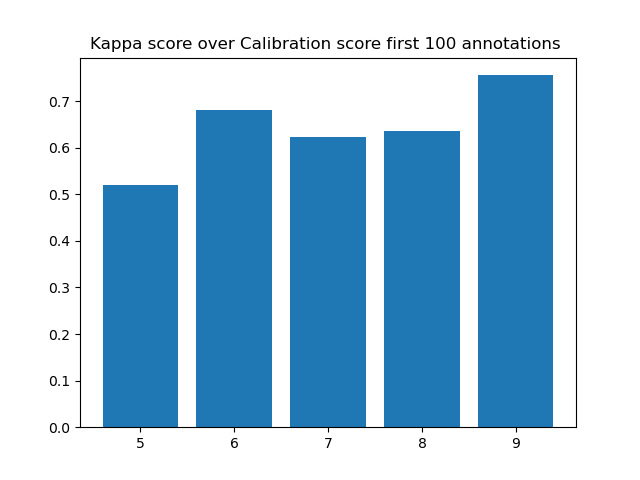}
		\caption{Kappa agreement $\kappa$ over calibration score for the first 100 full-scale 
		annotations. Shown x-values (calibration scores) are multiplied by ten.}
		\label{fig:qa-stats-2}
	\end{subfigure}
\end{figure}

In addition to the overall agreement, we measured the 
annotator agreement over evidence match ratios. As seen in Figure \ref{fig:qa-stats-1}, in the 
case of exact evidence match, the kappa agreement $\kappa$ is $0.92$, linearly decreasing with 
an agreement of $0.11$ in the case of completely distinct evidence. % It also has to be noted 
%that a larger proportion of annotations has a completely distinct evidence. This is mostly 
%attributed to searches that lead annotators to completely different pages and hence evidence. 

We further measured the annotation agreement sorted by annotator calibration score (more 
specifically the verdict accuracy) for the first 100 full-scale annotations of an annotator. As 
seen in Figure \ref{fig:qa-stats-2}, annotators with a calibration score of over $0.9$ have an 
overall higher kappa score, of around $0.8$ while annotators with a score of below $0.6$ only 
achieved a kappa score of about $0.5$. This indicates that the calibration score is indicative 
of the performance of annotators in the beginning. Looking at the score over all annotations, 
we however noticed, that annotators continue to align their annotations with annotators having 
very similar agreement irregardless of calibration score (except annotators with very 
calibration score above $0.9$ still having higher agreement).

\subsection {Dataset Processing \& Implementation Details }

\subsection{Dataset Processing}
Wikipedia articles were split into sentences using the NLTK unsupervised sentence 
tokenizer\footnote{\url{https://www.nltk.org/api/nltk.tokenize.html}}. We trained the 
unsupervised tokenizer on Wikipedia text to extract a large list of abbreviation words used on 
Wikipedia. These can be simply abbreviations of names (e.g. John F. Kennedy) or glossing 
abbreviations (for instance 'e.g.'). Due to the extensive use of Wikipedia templates for tables 
and the difficulty in resolving/parsing them, we opted in extracting articles from Wikipedia 
directly. We used Scrapy for this \footnote{https://scrapy.org/} and maintained a date stamp 
for each site. We limited the extracted tables to the classes 'wikitable' and 'infobox'. This 
restriction was set as there are contents of Wikipedia categorized as HTML tables while being 
highly specifically formatted, such as climate tables or tournament brackets. FEVEROUS 
maintains the diversity of Wikipedia tables/lists, only filtering ones out with formatting 
errors or that are empty (e.g. due to only containing images). %Similarly with listsWe only 
%filtered lists that have formatting errors or that are empty due to images or similar being 
%the 
%entire content of a list.

The FEVEROUS Wikipedia retrieval corpus was processed by keeping only hyperlinks with an 
associated article in the corpus. We replaced hyperlinks that are references to redirect pages 
with the respective page that the redirect page references to. URLS are replaced with a special 
token and text has been cleaned using the clean-text 
library\footnote{\url{https://pypi.org/project/clean-text/}}. 

For the annotation platform, we populated a MediaWiki 1.31 database with the extracted articles 
as well as Wikipedia redirects. We installed the CirrusSearch extension 
\footnote{\url{https://www.mediawiki.org/wiki/Extension:CirrusSearch}} to enable the search 
engine to use Elasticssearch as the back-end search. The annotations where stored in an SQL 
database using MariaDB.

\subsubsection{Implementation \& Evaluation Details}
\paragraph{Retriever.} We use Spacy\footnote{\url{https://spacy.io/}}, specifically the 
\texttt{en\_core\_web\_sm} model) to extract entities from claims. We match extracted entities 
against all titles of our Wikipedida database and extract pages with an exact match. The TF-IDF 
part of our retriever is largely based on DrQA \citep{chen2017reading}, computing the cosine 
similarity between the binned unigram and bigram TF-IDF vectors of claim  and the introductory 
section of a Wikipedia article. The same TF-IDF approach is used to extract sentences and 
tables, however, restricted to the top $k$ extracted pages. We excluded lists from the 
retrieval for our baseline to minimize computation time, considering that only 1\% of annotated 
evidence is located in lists.

The cell retrieval model uses pre-trained RoBERTa$_{base}$ from 
Huggingface\footnote{\url{https://huggingface.co/}}. Parts of the table that were longer than 
the maximum input length of RoBERTa were simply cut-off. To prevent this from happening during 
training we use row-sampling. We concatenate rows that contain relevant cell evidence first, 
before considering irrelevant rows. 

The cell retrieval RoBERTa classifier was fine-tuned using binary cross-entropy. The batch-size 
was set to $16$, with weight decay of $0.1$, a learning rate of $5e^{-5}$, and a total of $1$ 
training epochs. These hyperparameters are largely taken from recommendations 
\citep{devlin-etal-2019-bert} and have not been further fine-tuned as the baseline's purpose is 
not to achieve the highest possible scores, but rather to provide a working, intuitive model 
that motivates further exploration of the dataset. As such the models are not the main part of 
the paper. 

\paragraph{Verdict predictor}
The verdict predictor uses RoBERTa$_{large}$, particularly the model pre-trained on multiple 
NLI datsets by \citet{nie-etal-2020-adversarial}, which can be found 
\href{https://huggingface.co/ynie/roberta-large-snli_mnli_fever_anli_R1_R2_R3-nli}{here}. Each 
piece of evidence is separated using \texttt{</s>}. We linearize cell evidence the following : 
\texttt{[CONTEXT-HEADER] is [CELL]}, similar to \citep{schlichtkrull2020joint}. Our model is 
fine-tuned using a batch-size of $16$, a weight decay of $0.01$, a learning rate of  $1e^{-5}$ 
for $1$ epoch.  Similar to the cell retrieval model, these values are largely taken from 
reference and have not been fine-tuned. Same rationale here as stated above.

Experiments using RoBERTa have been repeated twice and the average was reported, with very low 
variance (around $2e^{-5}$). All experiments were done in Python3.7. We fine-tuned all models 
on a single Quadro RTX 8000. Fine-tuning the cell extractor took around 1.5h, while fine-tuning 
the verdict predictor took around 4h. The TF-IDF retrieval needed around 10h on a Xeon Gold 
5218 8 cores.

\subsection{Annotation details}

\begin{figure}[ht!]
	\centering
	\includegraphics[width=0.8\textwidth]{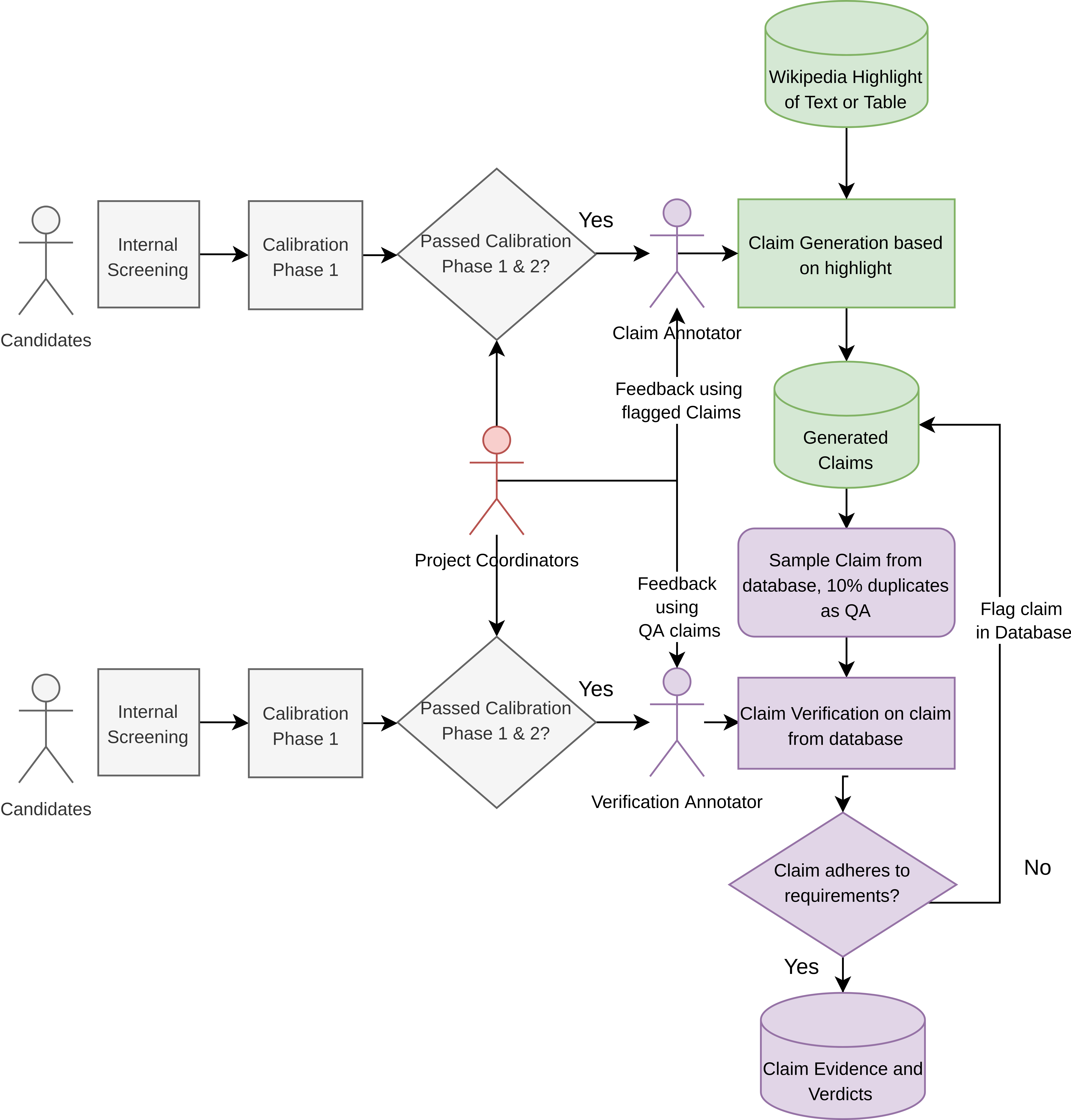}
	\caption{A schematic view on the annotation process of FEVEROUS. The claim generation 
	phase  is highlighted in green, the claim verification phase is noted in purple, and the 
	screening of annotators is highlighted in gray.}
	\label{fig:feverous-pipeline}
\end{figure}

The annotation process to create FEVEROUS is visualized in Figure \ref{fig:feverous-pipeline}.

\subsubsection{Annotation interfaces}
\paragraph{Navigation} To find relevant pages annotators can make use of the MediaWiki search 
functionality, a custom page search functionality, as well as hyperlinks. We aimed to create an 
ecosystem as realistic as possible, so annotators were motivated to approach this problem 
naturally: \emph{How would you search for relevant information to check the truthfulness of a 
statement/claim given to you?} The search bar shows relevant articles to annotator's search as 
soon as they start typing. They are further allowed to use the given recommendations and 
entering the main search page (i.e. clicking on 'Containing ...'). Tere are three kinds of 
hyperlinks: i) Hyperlinks embedded in a sentence, table or list, ii) Hyperlinks in the content 
box of each Wikipedia article, iii) Hyperlinks below section headers that refer to the main 
article or to a more specialized article.  Moreover, annotators had the option modify previous 
annotations.
\paragraph{{Operating the search engine}} The search engine allows annotators to simply type in 
words or phrases that they are looking for. If they type in a query into the engine it will 
show them suggestions if a \textbf{title matches the query}. If it cannot find a matching 
title, they can start a "full text search" i.e. \textbf{searching through the actual content of 
a page by clicking on ``\emph{containing...}" on the very bottom of the suggestions}. Doing so 
redirects them to a search page, with suggestions and highlights in articles where the query 
could be (partially) matched. While queries can simply be words or phrases, annotators could 
further modify their search queries with some operators (see Annotation Guidelines, however, 
these have been used only very rarely.

\begin{figure}[h]
	\centering
	\includegraphics[width=\textwidth]{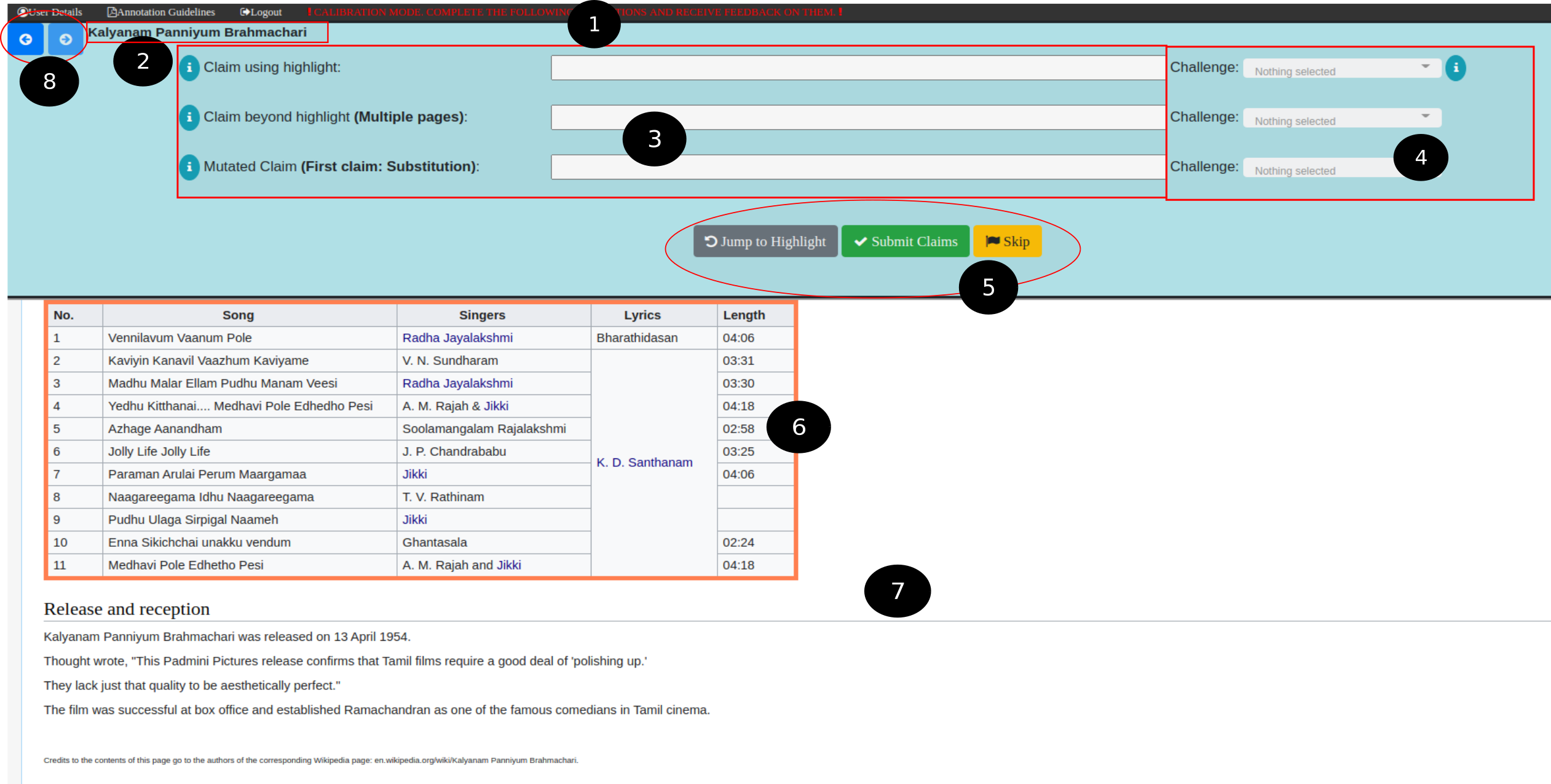}
	\caption[Interface]{Claim generation interface. \circledd{1} Menu bar with Access to User 
	Information, Annotation Guidelines, and Logout.  \circledd{2} Title of current Wikipedia 
	article.  \circledd{3} Text field for writing claims.  \circledd{4} Selection of expected 
	challenges.  \circledd{5} Buttons for i) Jumping back to the Wikipedia highlight, ii) 
	submitting the written claims and selected challenges, iii) skip the current highlight. 
	\circledd{6} Wikipedia article and navigation.  \circledd{7} Article highlight to base the 
	claims on. \circledd{8} Move between previously annotated claims. }
	\label{fig:interface_overview_ca}
\end{figure}

\begin{figure}[ht!]
	\centering
	\includegraphics[width=\textwidth]{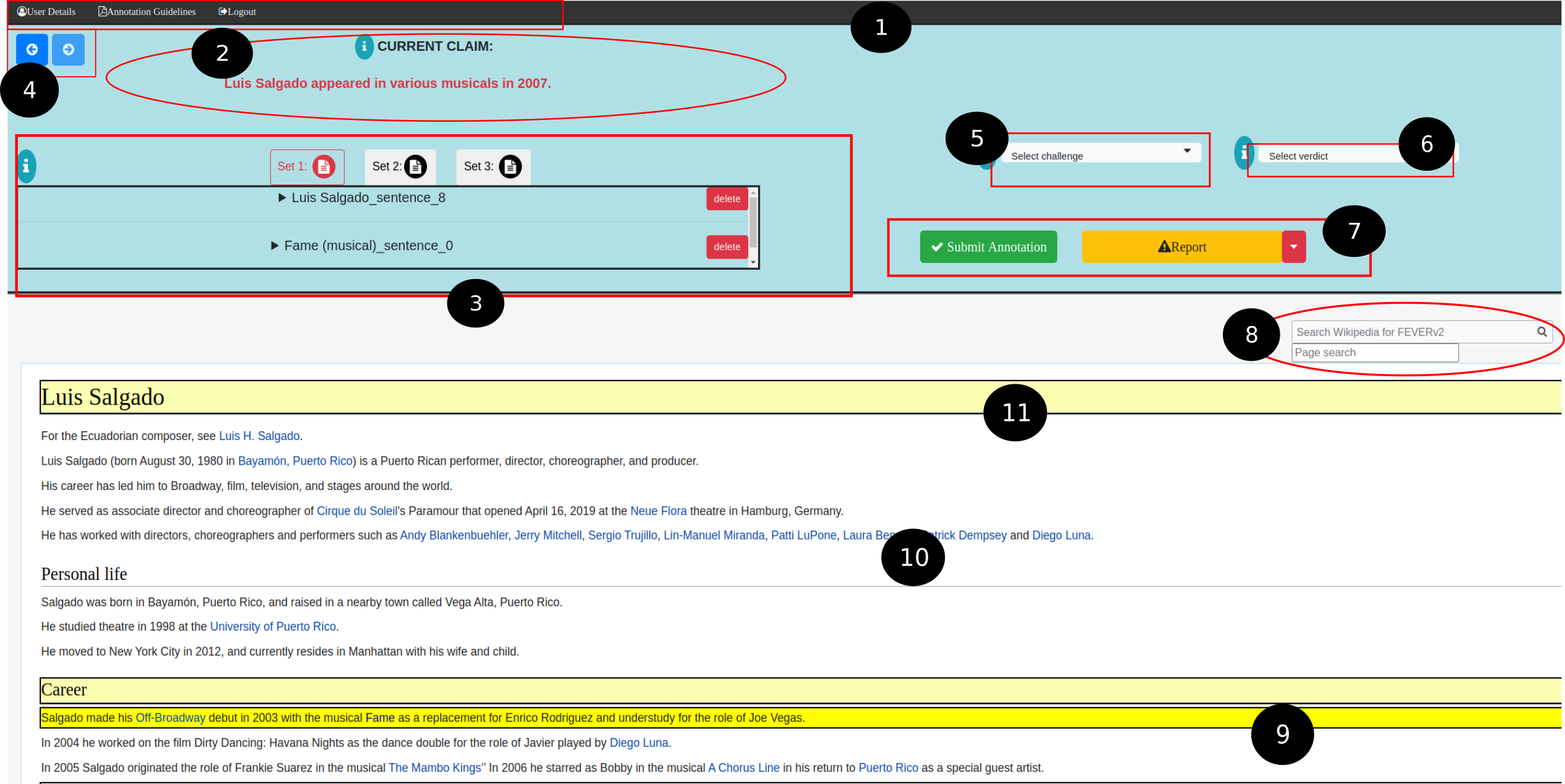}
	\caption[Interface]{claim verification Interface. \circledd{1} Menu bar with Access to User 
	Information, Annotation Guidelines, and Logout.  \circledd{2} Current claim to verify and 
	retrieve evidence for. \circledd{3} Move between previously annotated evidence. 
	\circledd{4} Management of selected evidence.  \circledd{5} Specifying annotation 
	challenges \circledd{6} Selection of the claims veracity (Supported, Refuted, 
	NotEnoughInformation) \circledd{7} Button for submitting annotating/reporting claim 
	\circledd{8} Search bars for i) navigating through Wikipedia articles, ii) information 
	filtering within a Wikipedia page. \circledd{9} WikiMedia interface \circledd{10} Selected 
	evidence (yellow highlighting) \circledd{11} Corresponding evidence context (lighter yellow 
	highlighting).}
	\label{fig:interface_overview_ea}
\end{figure}

\subsection{Claim generation}
\label{sec:claim-generation}

\subsubsection{Guidelines}
\paragraph{Generating Claim using highlight (Type I)}
The first claim should \textbf{exclusively use information from the highlighted 
table/sentences}. Only the page title and/or section title the highlight is located might be 
used for the claim as well. The claim must either align with the contents in the highlight or 
contradict them, indicated on the tool (i.e. true and false claims). A claim should adhere to 
following requirements:

\begin{itemize}
	\item A claim based on a table highlight should combine information of multiple cells if 
	possible. This includes comparisons (e.g. \emph{X scored higher/lower than Y}, or 
	\emph{While X was the son of Z, Y was the son of Q.}), superlatives (\emph{X scored the 
	highest/lowest}, or \emph{X was the first Japanese supercomputer. }), filters (\emph{X, Y 
	and Z scored more than 10 points}, or \emph{X, Y, Z are manufactured in Germany.}), and 
	arithmetic operations (\emph{5 teams scored more than 10 points}, or \emph{X was born 2 
	years and 8 months before Y.}). % Add example of non-numerical data. Use examples from 
	%previous annotaionas
	
	\item A claim based on highlighted sentences should not simply paraphrase a highlighted 
	sentence or concatenate sentences. Instead, information of multiple sentences must be 
	combined. Information from at least two sentences must be used for generating the claim.  
	
	\item A claim should be a single well-formed sentence. It should end with a period; it 
	should follow correct capitalization of entity names (e.g. `India', not `india'); numbers 
	can be formatted in any appropriate English format (including as words for smaller 
	quantities).
	
	\item Generated claims must \textbf{\underline{not be subjective}} and be 
	\textbf{\underline{verifiable}} using publicly available information/knowledge.\\
	\textit{Don't}: John Lennon was a more popular musician than Tommy Moore.\\
	\textit{Do}: John Lennon's discography sold two times as many box sets as Tommy Moore in 
	1997.
	\textit{Don't}: Sea Songs by Yadollah Royaee (born in 1932) is rich in symbolism and is 
	deeply inspired by Persian mysticism. 
	\textit{Do}: Sea Songs by Yadollah Royaee (born in 1932) contains symbolism and is inspired 
	by Persian mysticism. 
	
	\item The claim should be \textbf{as unambiguous as possible} and avoid vague or 
	speculative language (e.g. might be, may be, could be, rarely, many, barely or other 
	indeterminate count words)\\%in terms of the claim itself as well as information
	\textit{Don't}: The Olympic Games have rarely taken places in Europe\\
	\textit{Do}: The Olympic Games were held three three times in Europe.
	\textit{Don't}: Michael Ballack scored the most goals.\\
	\textit{Do}: Michael Ballack scored the most goals in the Bundesliga 2004/2005 season.
	
	\item A claim must not contain any idioms, figures of speech, similes, or verbose language.
	\textit{Don't}: The scientist Mary Lamb owned five sheep with fleece as black as coal, but 
	they were not used in any of her experiments.\\
	\textit{Do}: The scientist Mary Lamb owned five sheep with black fleece, but they were not 
	used in any of her experiments.
	
	\item The claim must be understood by itself (i.e. no pronouns) --  \emph{[Note: in the 
	case where the highlighted text does not contain a mention of the entity at question, you 
	should use the title of the page or the header of the section for that information]}.\\
	\textit{Don't}: He played most of his football career for Chelsea.\\
	\textit{Do}: Didier Drogba played most of his football career for Chelsea.\\

	\item Claims should not be about contemporary political topics  (e.g. contemporary Wars 
	(from the second world war and onwards), disputed topics) -- skip pages where the 
	highlighted area only discusses such topics.\\
	Don't: In 1974 Turkey had landed 30,000 troops on Cyprus and captured Kyrenia.
	
	\item  In some cases highlighted Wikipedia information is not correct/consistent. These 
	highlights are still valid for claim generation. For this workflow don't worry about the 
	factual correctness of Wikipedia. If you think that the highlighted information is 
	disputed, better skip it.
	
	\item Do not incorporate your own knowledge, believes or additional world knowledge into 
	the claim. Focus only on the highlighted Wikipedia section given to you!			
\end{itemize}

\paragraph{Generating Claim beyond the highlight (Type II)}

The second claim should be based on the highlight, but \textbf{must include information beyond 
the highlighted table/sentences}. You are free in deciding to modify the previously created 
claim that uses only the highlight or to create an unrelated one (that still includes 
information from the highlight). Either way, the new claim must still adhere to the 
requirements mentioned above. The new claim can either be supported or refuted. So in general, 
you should not worry whether the new claim preserves the truth value of the first claim. 
However, please keep in mind that we aim for having a similar number of positive vs negative 
claims. 
Information to include must either be on the same page or from other Wikipedia pages, indicated 
on the tool:

\begin{enumerate}
	\item \textbf{Same page:} Include information outside of the highlight but on the same page.
	\item \textbf{Multiple pages:} Include information from other Wikipedia page(s). You can 
	search freely through Wikipedia using the search function, available hyperlinks on the 
	pages, and the \emph{Return to highlight} button.
\end{enumerate}

Moreover, for this claim it is allowed to use information/knowledge that might not be available 
in Wikipedia but you assume to be general knowledge, e.g.\ 
that 90s refers to the timespan from 1990 to 1999. Similarly to the previous claim, the claim 
can either align with the used information or contradict it.
We encourage you to create claims that are based on a combination of structured and 
unstructured information: tables, sentences, lists, captions, or section titles.

Example 1:\\
\textit{Claim using highlight}: The Zuse Z3 was program-controlled by punched 35mm film stock.\\
\textit{Claim using more than highlight}: Programs were executed on the Zuse Z3 by using 
punched 35mm film stock with manually entered initial values. \\

Example 2:\\
\textit{Claim using highlight}: The player with the most number of total assists at Shrewsbury 
Town F.C in 2013 is Luke Summerfield.\\
\textit{Claim using more than highlight}: The player with the most number of total assists at 
Shrewsbury Town F.C in 2013 also played
for Liverpool.

Example 3:\\
\textit{Claim using highlight}: Jeff Gordon had the most points at the 1998 Pepsi 400 stock car 
race.\\
\textit{Claim using more than highlight}: Jeff Gordon's points at the end of the Winston cup in 
1998 were higher than the points of all drivers at 1998 NAPA 500 combined. \\

\paragraph{Mutated Claim (Type III)}
We additionally ask the annotators to modify one of the two claims with one of the following 
\emph{mutations types}: \textbf{More Specific, Generalization, Negation, Paraphrasing, Entity 
Substitution, Tense Shift}. Both the type of modification and which of the two claims to be 
modified are specified in the interface (see \circledd{3} in Figure 
\ref{fig:interface_overview_ca}). Similar to 'Claim beyond highlights', the modification can 
result in a claim that can either be supported or refuted. So in general, you should not worry 
whether the mutation will preserve the truth of the claim or not. Again, for this claim it is 
allowed to use information/knowledge that might not be available in Wikipedia but you assume to 
be general knowledge. Make sure that the new claim is still a single sentence! Here is an 
explanation for each mutation type:

\begin{enumerate}
	\item \textbf{Generalization} Make the claim more general so that the new claim is a 
	generalization of the original claim (by making the meaning less specific)
	\item \textbf{More Specific} Make the claim more specific so that the new claim
	is a specialization (as opposed to a generalization) of
	the original claim (by making the meaning more specific). 
	\item \textbf{Negation} Negate the meaning of the claim. This is not to be confused with 
	making claim false: negating the meaning of a claim could make a false claim true and vice 
	versa!
	\item \textbf{Paraphrasing} Rephrase the claim so that it has the same meaning
	\item \textbf{Entity Substitution} Substitute an entity in the claim to alternative from 
	either the same or a different set of things. If the object in the claim is an entity, 
	replace this entity. Chose any entity in the claim otherwise. 
\end{enumerate}

Given the claim "\emph{John E. Moss was a politician of the US Democratic party.}" Table 
\ref{tab:manipulation}  shows each modification for the example sentence, following table 
shows  example modifications for each mutation type: 

\begin{table}[ht!]
	\begin{tabulary}{1\textwidth}{C R}
		\textbf{Type} & \textbf{Modified Claim}\\
		\hline
		More specific & John E. Moss was a politician of the US Democratic party for 
		California's 3rd congressional district. \\
		\hline
		Negation & John E. Moss has never ran for office. \\
		\hline
		Generalization & John E. Moss was a US American politician. \\
		\hline
		Paraphrase  & John E. Moss was a US American politician of the Democratic party. \\
		\hline
		Entity Substitution & John E. Moss was a politician of the US Republican Party. \\
		\hline
		
	\end{tabulary}
	\caption{Claim manipulation for the claim "\emph{John E. Moss was a politician of the US 
	Democratic party.}"}
	\label{tab:manipulation}
\end{table}

\paragraph{Expected Main Verification Challenge}
\label{sec:challenge_anno}
We want to know what you think is the main challenge for assessing the veracity and retrieving 
evidence for the claim you have created. You must select one of the given challenge categories 
you expect to be the main challenge: \textbf{Multi-hop Reasoning, Numerical Reasoning, 
Combining Text and Tables, Entity Disambiguation}, and \textbf{Search terms not in claim}. If 
the main challenge hasn't been any of these, select \textbf{Other}.

\begin{enumerate}
	\item \textbf{Multi-hop Reasoning}  Multi-hop reasoning expected to be the main challenge 
	for verifying that claim, i.e. several pages/sections will be required for verification. 
	e.g. "\emph{The player who ranked 3rd at the US Open in 2010 played in the most populated 
	city of Germany in 2014}".
	\item \textbf{Numerical Reasoning} Numerical reasoning expected to be the main challenge to 
	verify the claim, i.e. reasoning that involves numbers or arithmetic calculations. This 
	also includes steps such as counting cells in tables. Example: Given a claim "A is older 
	than B", and for both A and B only their birth dates are given, concluding the older person 
	would require mathematical inference. Another example would be given the following scores 
	in tennis '7-4', 2-6', and 6-1' to conclude that Player 1 won the match.
	% 	\item \textbf{Reasoning over Structure} Reasoning over Table or List(s) is required and 
	%it posed a challenge due to complex reasoning on tables, due to combining list(s)/table(s) 
	%with text, and/or due to combining or counting several pieces of information (e.g. In a 
	%table with people, birth-dates, year of death, and occupation, to find the engineer who 
	%lived the longest).
	\item \textbf{Combining Tables and Text} Combining list(s)/table(s) with information from 
	text (i.e. phrases, captions, sentences) outside tables is expected to be the main 
	challenge, i.e. when the Text provides important context to Tables/List to be understood 
	and vice versa (titles are excluded when talking about text in this challenge).
	\item \textbf{Entity disambiguation} Disambiguating an entity is expected to be the main 
	challenge for verifying a given claim. E.g. Adam Smith was a footballer for the Bristol 
	Rovers (Wikipedia lists 4 Adam Smiths that played football).
	\item \textbf{Search terms not in claim}
	The main challenge is expected to be finding relevant search terms to pages with required 
	evidence to verify a given claim goes beyond searching for terms located in the claim 
	itself, e.g. for the Claim "\emph{Non college educated voters voted 67 percent for the 
	democratic party in 1952}" the evidence is located on the page "New Deal Coalition" -- 
	challenging to deduce the page based on the claim. Evidence that can quickly be found by 
	searching for an entity mentioned in the claim is most likely not a retrieval challenge 
	(excluding entity mentions that could refer to many entities).  
	
	\item \textbf{Other} If none of the above challenges can be identified. %If unsure whether 
	%to select \emph{Other} or another label, always select the latter.
\end{enumerate}		
\subsubsection{Examples}
\begin{figure}[ht!]
	\centering
	\includegraphics[width=\textwidth]{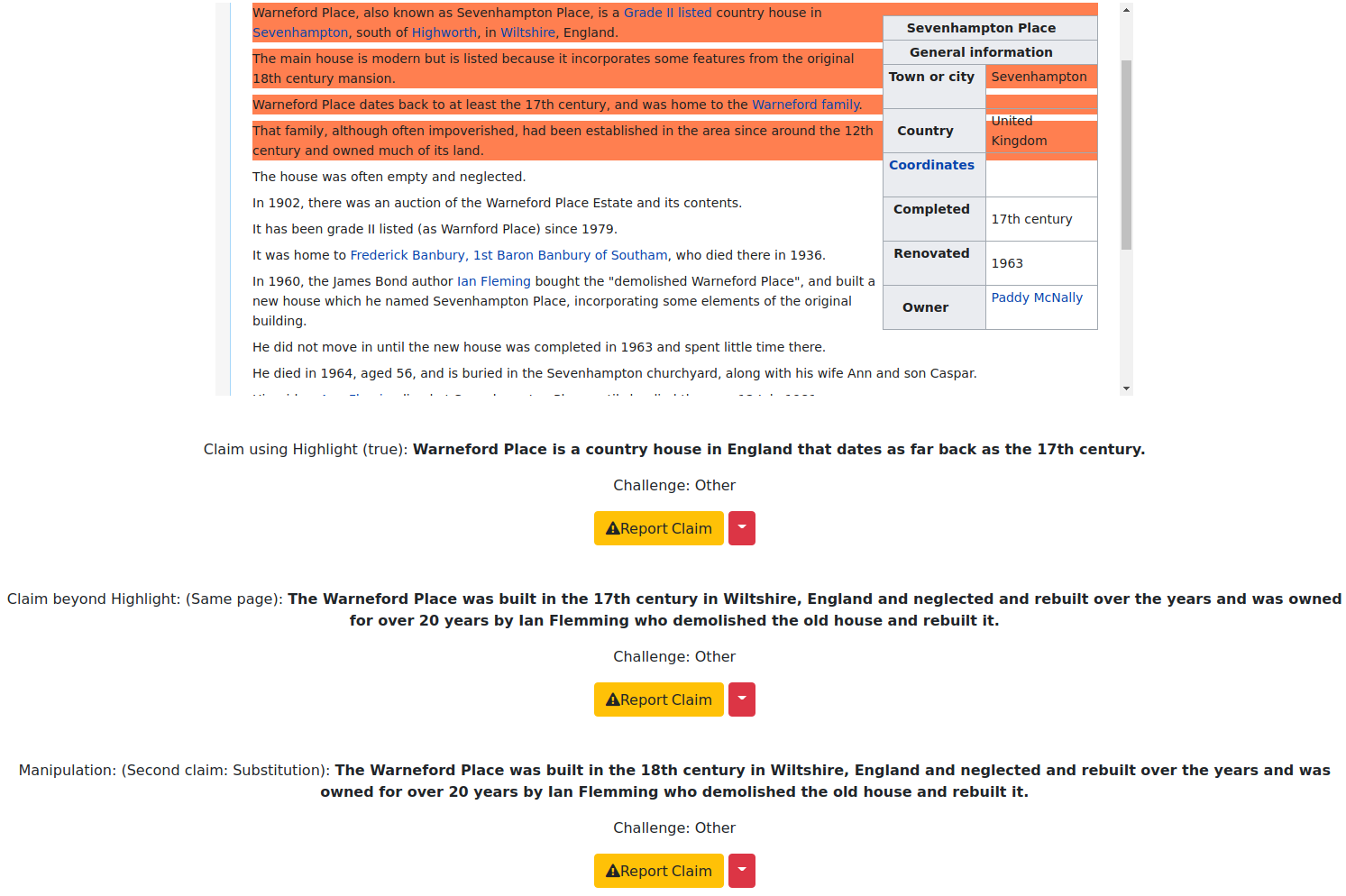}
	\caption{Example claim generation annotation, given a sentence highlight.}
	\label{fig:claim-generation-ex1}
\end{figure}

\begin{figure}[ht!]
	\centering
	\includegraphics[width=\textwidth]{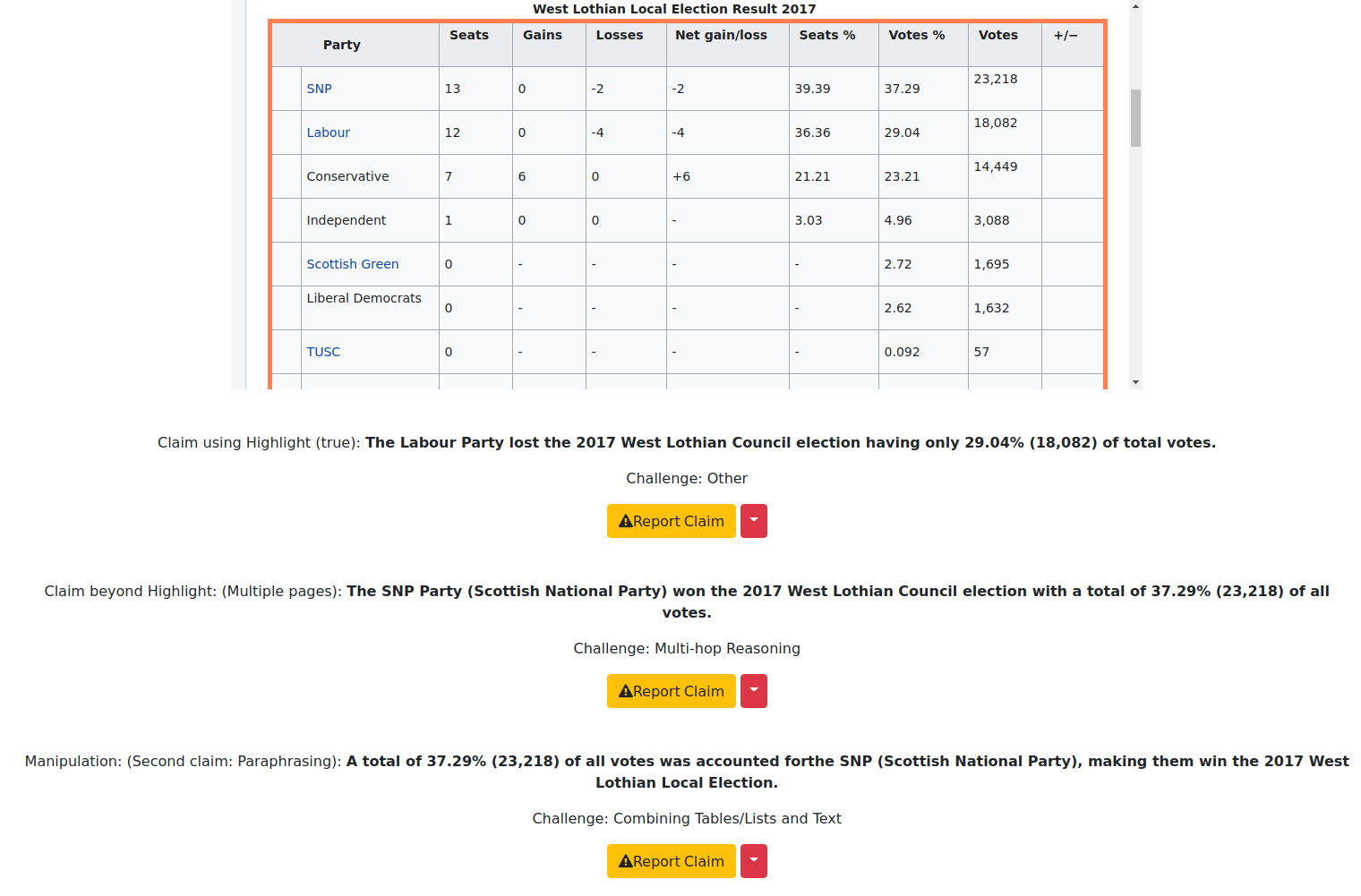}
	\caption{Example claim generation annotation, given a table highlight.}
	\label{fig:claim-generation-ex2}
\end{figure}

See Table \ref{fig:claim-generation-ex1} and \ref{fig:claim-generation-ex2} for examples.

\subsection{Claim Verification}

\subsubsection{Guidelines}
\paragraph{Evidence highlighting}
As soon as relevant information has been found in either \emph{text (sentences or table 
captions), tables, or lists} you can add it as evidence to your annotation by clicking on it. 
For free text, the \textbf{entire sentence/phrase} will be selected as evidence. For tables, 
one \textbf{cell} is selected, and finally for lists, one \textbf{item} will be highlighted. 
Evidence from different Wikipedia can be freely combined -- there are no restrictions. There is 
also no limitation in terms of evidence pieces required to validate a claim. However, an entire 
annotation for a single claim should not surpass \textbf{10 Minutes}. If it does, keep the 
already annotated evidence and submit it with the verdict  \emph{NotEnoughInformation}.\\

For every highlighted sentence/cell/list item some context is extracted automatically and shown 
to you in the interface. Article titles and sections (and subsections, subsubsections etc.) in 
which the evidence is located are always extracted. Additionally, if a cell has been 
highlighted the corresponding table headers are extracted as well. Due to the complexity and 
diversity in Wikipedia tables, it is possible that some additional table headers have not been 
highlighted automatically, but would still be needed to interpret the selected evidence 
correctly.
%that still add information to the content or direct context of that cell.
\textbf{These headers need to be highlighted manually by you.}\\

You must apply common-sense reasoning to the evidence you read but avoid applying your own 
(world) knowledge. If possible, additional evidence should be highlighted which provides the 
missing information (e.g. that a \emph{Democrat} is a politician of the Democratic Party, or 
that \emph{60s} refers to the years 1960 -- 1969). If this very general world knowledge cannot 
be found on Wikipedia you are nonetheless allowed to use it for the verdict or to find further 
evidence.  Be careful that you do not use your knowledge to reach hasty conclusions, for 
instance given the evidence X is goalkeeper and the captain of Team Y, we do not have enough 
support that 'X is the starting goalkeeper for Y'. While it is often the case that the first 
implies the second, it is not always true.\\

As a guide - you should ask yourself: \textbf{\emph{If I was given only the selected sentences, 
table cells, and list items shown in the evidence overview \circledd{3}, do I have strong 
enough reason to believe the claim is supported or strong enough reason to believe the claim is 
refuted. If I’m not certain, what additional information do I have to add to reach this 
conclusion and can I find it on Wikipedia?}}\\
While claims that are \emph{Supported} require evidence for each fact mentioned in that claim 
as far as possible, \emph{Refuted} claims must only select evidence of the information that 
contradicts (parts of) the claim. If a refuted claim is partially supported, do not provide 
evidence for the partially supporting parts, unless it is necessary context for the refuting 
evidence, e.g.\ ensuring the correct entity is being referred to. If a claim is marked as 
\emph{NotEnoughInformation} please still submit the evidence found in the process of reaching 
this verdict!\\

All your annotated evidence (excluding titles, and sections, but including table headers) is 
shown to you in the evidence overview \circledd{3}, Figure \ref{fig:interface_overview_ea}. You 
initially see the ID of the annotated evidence, however, by clicking on the ID in the overview 
it will expand and show you the actual content of the element you selected. This way you can 
keep track of evidence from possibly multiple pages easily. If you change your mind and want to 
remove a piece of evidence simply click again on the now highlighted element.

\begin{titlemize}{ \color{orange}  Note!}
	\item If the verification of a claim requires to include every entry in a table row/column 
	(e.g. claims with universal quantification such as `highest number of gold medals out of 
	all countries'), you must highlight each cell of that row/column (c.f. Example 4, 7).
	\item Content on Wikipedia that contains qualifier or hedges (e.g. probably, likely, might) 
	should not be used as evidence. For instance, a sentence such as \emph{Michael Mueller was 
	likely not involved in the 2012 scandal} should not be considered as evidence for the given 
	claim.
	\item If you are not able to find any evidence for the given claim, you are still required 
	to submit the annotation. As mentioned above, select the verdict \emph{Not enough 
	Information} in this case and challenges (as described below)
	\item Make sure that you find evidence to support each fact mentioned in a claim when 
	selecting Supported, especially for longer claims. For instance given the claim "The 
	scientist Mary Lamb owned five sheep with black fleece, but they were not used in any of 
	her experiments.", the claim can be broken down into five pieces of information that all 
	need to be verified in order to select supported: \\
	1. There exists a person named Mary Lamb\\
	2. Mary Lamb is a scientist\\
	3. Mary Lamb owned five sheep\\
	4. Those sheep had black fleece\\
	5. The sheep that Mary owned were not used in any of her experiments
	\item Do not take possible motives of Wikipedia editors into account when assessing the 
	evidence -- take the evidence as it is.
	\item When highlighting cells in very large tables there could be a delay until the cell 
	and the automated context are highlighted. This is because the table has to be processed 
	before the correct context is identified.
	\item There exists no interaction between different claim verification annotators in the 
	interface -- do not worry about this!
	\item Even if entire sentences are located in tables or lists, the finest granularity 
	remains the cell or item, respectively. Therefore, the entire content of the cell will be 
	added which is fine!
\end{titlemize}

\paragraph{Ambigious \& Misleading Claims}
In cases where you could find multiple ways of interpreting the claim which give rise to 
different verdicts, ask yourself the following question: \emph{\textbf{Would you consider 
yourself misled by the claim given the evidence you found?}} 
For instance, take the claim "\emph{Shakira is Canadian}". Even if the evidence only concludes 
that she is Colombian (not a direct contradiction to the claim), it is still okay to refute the 
claim as there is enough evidence to believe that the claim is misleading, according to common 
perception.  Similar case with a claim "Lamb owned five sheep", given the sentence  "Lamb had a 
love of farming and owned many barnyard animals, including two hens and four sheep", we can 
conclude that the claim is misleading and thus refuted.\\

If you have doubt regarding your assessment go with NEI, e.g. given the claim 'Shakira was 
diagnosed with Diabetes Type II' and the evidence that Shakira was diagnosed with Diabetes when 
she was 10, it is clear to someone with specialized knowledge that the claim is false, however 
as it goes beyond common perception it is NEI. Do not include any knowledge about how the 
claims are generated when evaluating how misleading a claim is, e.g.\ that this claim is likely 
to be a corrupted version of the claim ``\emph{Shakira is Colombian}''.

\paragraph{Reporting a claim}
It is possible to report and skip a given claim. It might be appropriate to flag a claim if i) 
the claim is personal, implausible, not verifiable, not understandable by itself, or too 
ambiguous ii) does not meet other aspects of the guidelines from the Claim generation task 
(i.e. not containing idioms, figures of speech, similes, verbose language, and not be about 
contemporary political topics)*, iii) ungrammatical claims or typographical errors, spelling 
mistakes iv) required evidence is not displayed correctly. When reporting a claim select the 
appropriate action from the menu or write an individual text. Do not skip a claim if it is 
phrased similarly to another one you have already annotated. We explicitly include paraphrased 
claims for annotation as we want to gather claim verifications for these too.\\

\paragraph{Main Verification Challenge}
\label{sec:challenge}
We are interested in gaining more insights into the main challenge the annotator had for 
finding evidence for the given claim. You must select one of the given challenge categories: 
\textbf{Multi-hop Reasoning, Numerical Reasoning, Combining Tables and Text, Entity 
Disambiguation}, and \textbf{Search terms not in claim}. If the main challenge hasn't been any 
of these, select \textbf{Other}. 

\begin{enumerate}
	\item \textbf{Multi-hop Reasoning}  Multi-hop reasoning was the main challenge challenge 
	for verifying that claim, i.e. several pages/sections will be required for verification. 
	e.g. "\emph{The player who ranked 3rd at the US Open in 2010 played in the most populated 
	city of Germany in 2014}".
	\item \textbf{Numerical Reasoning} Numerical reasoning was the main challenge when 
	verifying the claim, i.e. reasoning that involves numbers or arithmetic calculations.  This 
	also includes steps such as counting cells in tables. Example: Given a claim "A is older 
	than B", and for both A and B only their birth dates are given, concluding the older person 
	would require mathematical inference. Another example would be given the following scores 
	in tennis '7-4', 2-6', and 6-1' to conclude that Player 1 won the match.
	\item \textbf{Combining Tables and Text} Combining list(s)/table(s) with information from 
	text (i.e. phrases, captions, sentences) outside tables was the the main challenge, i.e. 
	when the Text provides important context to Tables/List to be understood and vice versa 
	(titles and sections are excluded when talking about text in this challenge).
	\item \textbf{Entity disambiguation} Disambiguating an entity was the main challenge for 
	verifying a given claim. E.g. Adam Smith was a footballer for the Bristol Rovers (Wikipedia 
	lists 4 Adam Smiths that played football).
	\item \textbf{Search terms not in claim}
	The main challenge was finding relevant search terms to pages with required evidence to 
	verify a given claim goes beyond searching for terms located in the claim itself, e.g. for 
	the Claim "\emph{Non college educated voters voted 67 percent for the democratic party in 
	1952}" the evidence is located on the page "New Deal Coalition" -- challenging to deduce 
	the page based on the claim. Evidence that can quickly be found by searching for an entity 
	mentioned in the claim is most likely not a retrieval challenge (excluding entity mentions 
	that could refer to many entities).  
	
	\item \textbf{Other} If none of the above challenges can be identified.
\end{enumerate}

\subsubsection{Examples}
In addition to Figure 1 in the main paper, two further examples are shown in Figure 
\ref{fig:feverous_examples_2}.

\begin{figure}
	\begin{tabular}{ cc }
		\fbox{\begin{minipage}{19em}
				\small    
				\textbf{Claim:} Mike Ledwith (a professional baseball player) played one game 
				in MLB and scored one run.\\
				\rule{\linewidth}{0.1em}
				\centering{\textbf{Evidence}} \\
				\textbf{P:} \texttt{wiki/Mike Ledwith}\\
				\textbf{S0:} \texttt{Introduction}\\
				\textbf{e$_1$:} Michael Ledwith, was a professional baseball player who played 
				catcher in one game for the 1874 Brooklyn Atlantics. \\
				\textbf{S0:} \texttt{Introduction}\\
				\textbf{e$_2$:} \centering{
					\begin{tabular}{|r r|}
						\toprule
						\multicolumn{2}{|c|}{\cellcolor{gray!25}{Mike Ledwith}} \\
						\midrule
						\multicolumn{2}{|c|}{\cellcolor{gray!25}{MLB statistics}} \\
						\midrule
						\cellcolor{gray!25}{Games played} & \cellcolor{red!25}{1} \\
						\cellcolor{gray!25}{Runs} scored & \cellcolor{red!25}{1}\\
						\cellcolor{gray!25}{Hits} & 1 \\
						\cellcolor{gray!25}{Batting average} & 0.250\\
						\bottomrule
					\end{tabular}
				}
				\rule{\linewidth}{0.1em}
				\textbf{Verdict:} Supported\\
				\textbf{Expected Challenge:} Combining Tables and Text\\
				\textbf{Challenge:} Combining Tables and Text\\
		\end{minipage}} &

		\fbox{\begin{minipage}{19em}
				\small    
				\textbf{Claim:} Braeden Lemasters, an American actor, musician, and voice 
				actor, appeared in six films since 2008 and also appeared in TV shows such as 
				Six Feet Under where he starred as Frankie.\\
				\rule{\linewidth}{0.1em}
				\centering{\textbf{Evidence}} \\
				\textbf{P}: \texttt{wiki/Braeden\_Lemasters}\\
				\textbf{S2}: \texttt{Filmography}\\
				\textbf{e$_1$:} Braeden Lemasters (born January 27, 1996) is an American actor, 
				musician, and voice actor. \\
				\textbf{e$_2$}: \centering{
					\begin{tabular}{|r r r|}
						\toprule
						\cellcolor{gray!25}{Year} & \cellcolor{gray!25}{Film} & 
						\cellcolor{gray!25}{Role} \\
						\midrule
						2008 &  \cellcolor{red!25}{Beautiful Loser} & Jake \\
						2009 & \cellcolor{red!25}{The Stepfather}  & Sean Harding	\\
						2010 & \cellcolor{red!25}{Easy A} & 8th Grade Todd	 \\
						2012 &  \cellcolor{red!25}{A Christmas Story} 2& Ralphie Parker \\
						2017 &  \cellcolor{red!25}{Totem} & Todd\\
						2017 & \cellcolor{red!25}{Flock of Four} &  Joey Grover \\
						\bottomrule
					\end{tabular}
				}
				\textbf{S1}: \texttt{Life and career}\\
				\textbf{e$_3$:} In 2005, Braeden started his career at age 9, as Frankie, on 
				the TV show Six Feet Under. \\
				\rule{\linewidth}{0.1em}
				\textbf{Verdict:} Supported\\
				\textbf{Expected Challenge:} Combining Tables and Text\\
				\textbf{Challenge}: Combining Tables and Text\\
		\end{minipage}}
	\end{tabular}
	
	\caption{Two examples from the FEVEROUS dataset that require both unstructured and 
	structured information. The dataset contains both short, simple claims (left) and complex 
	claims (right).}
	\label{fig:feverous_examples_2}
\end{figure}

\subsubsection{QA annotation interface}
QA data was also used to recognise guidelines aspects that needed further clarification. 
Clarifications were communicated through updated guidelines as well as multiple FAQs. QA 
annotations were also used on an individual annotator level in combination with production 
reports, which measured statistics such as the number of claims an annotator generated that 
have been reported by verification annotators, to identify error patterns and giving annotators 
further individual feedback. An interface was provided to annotators to see the annotations 
that have been quality checked and to allow them to maintain an overview on their performance. %

\begin{figure}[ht!]
	\centering
	\includegraphics[width=\textwidth]{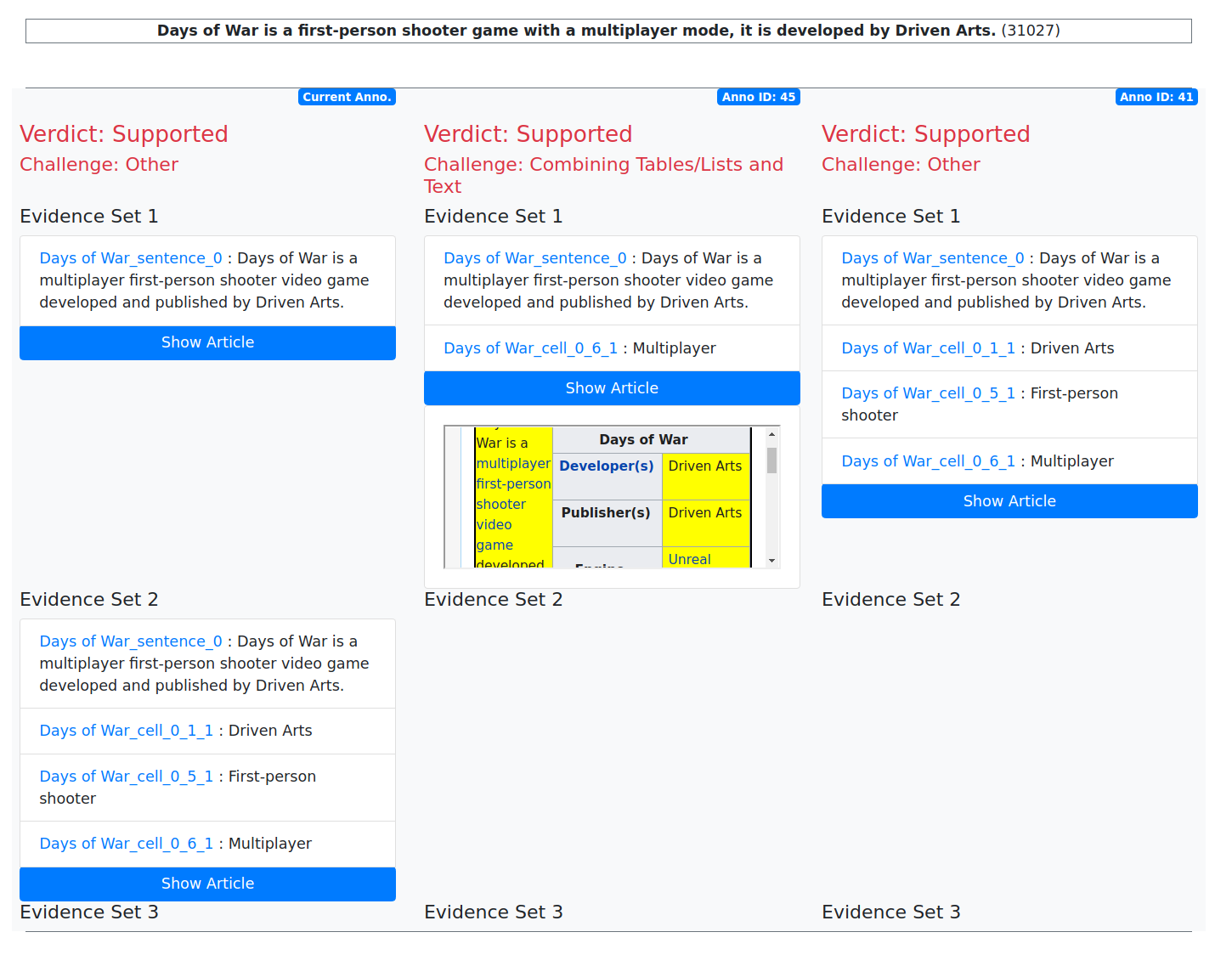}
	\caption{QA annotation interface for project managers. Interface for annotators looks 
	similar, with ID's being anonymized}
	\label{fig:qa-interface}
\end{figure}

Figure \ref{fig:qa-interface} shows the QA interface for project managers. QA annotations with 
only partial agreement or complete disagreement are highlighted in the interface in red. The QA 
interface for annotators looks similar, with ID's being anonymized.

\subsection{Author statement}
The authors of this paper bear all responsibility in case of violation of copyrights associated 
with the FEVEROUS dataset.

\end{document}